\documentclass[10pt,twocolumn,letterpaper]{article}

\pdfoutput=1

\usepackage{cvpr}
\usepackage{times}
\usepackage{epsfig}
\usepackage{graphicx}
\usepackage{amsmath}
\usepackage{amssymb}

\DeclareMathOperator*{\argmin}{arg\,min}

\usepackage[pagebackref=true,breaklinks=true,letterpaper=true,colorlinks,bookmarks=false]{hyperref}

\cvprfinalcopy 

\ifcvprfinal\fi
\begin{document}

\title{Virtual View Networks for Object Reconstruction}

\author{Jo\~{a}o Carreira, Abhishek Kar, Shubham Tulsiani and Jitendra Malik\\
University of California, Berkeley - Berkeley, CA 94720\\
{\tt\small \{carreira,akar,shubtuls,malik\}@eecs.berkeley.edu}
}

\maketitle

\begin{abstract}

All that structure from motion algorithms ``see'' are sets of 2D points. We show that these impoverished views of the world can be faked for the purpose of reconstructing objects in challenging settings, such as from a single image, or from a few ones far apart, by recognizing the object and getting help from a collection of images of other objects from the same class. We synthesize virtual views by computing geodesics on novel networks connecting objects with similar viewpoints, and introduce techniques to increase the specificity and robustness of factorization-based object reconstruction in this setting. We report accurate object shape reconstruction from a single image on challenging PASCAL VOC data, which suggests that the current domain of applications of rigid structure-from-motion techniques may be significantly extended.
\end{abstract}

\section{Introduction}

Modern structure from motion (SfM) and multiview stereo approaches \cite{snavely2006photo,crandall2011discrete,furukawa2010accurate} are widely used to recover viewpoint and shape information of objects and scenes in realistic settings, but require multiple images with overlapping fields of view. If only a single image of the target object is available, or if multiple ones are available but from viewpoints far apart, these methods are, respectively, inapplicable or prone to fail.

\begin{figure}[t]
\includegraphics[width=1\linewidth]{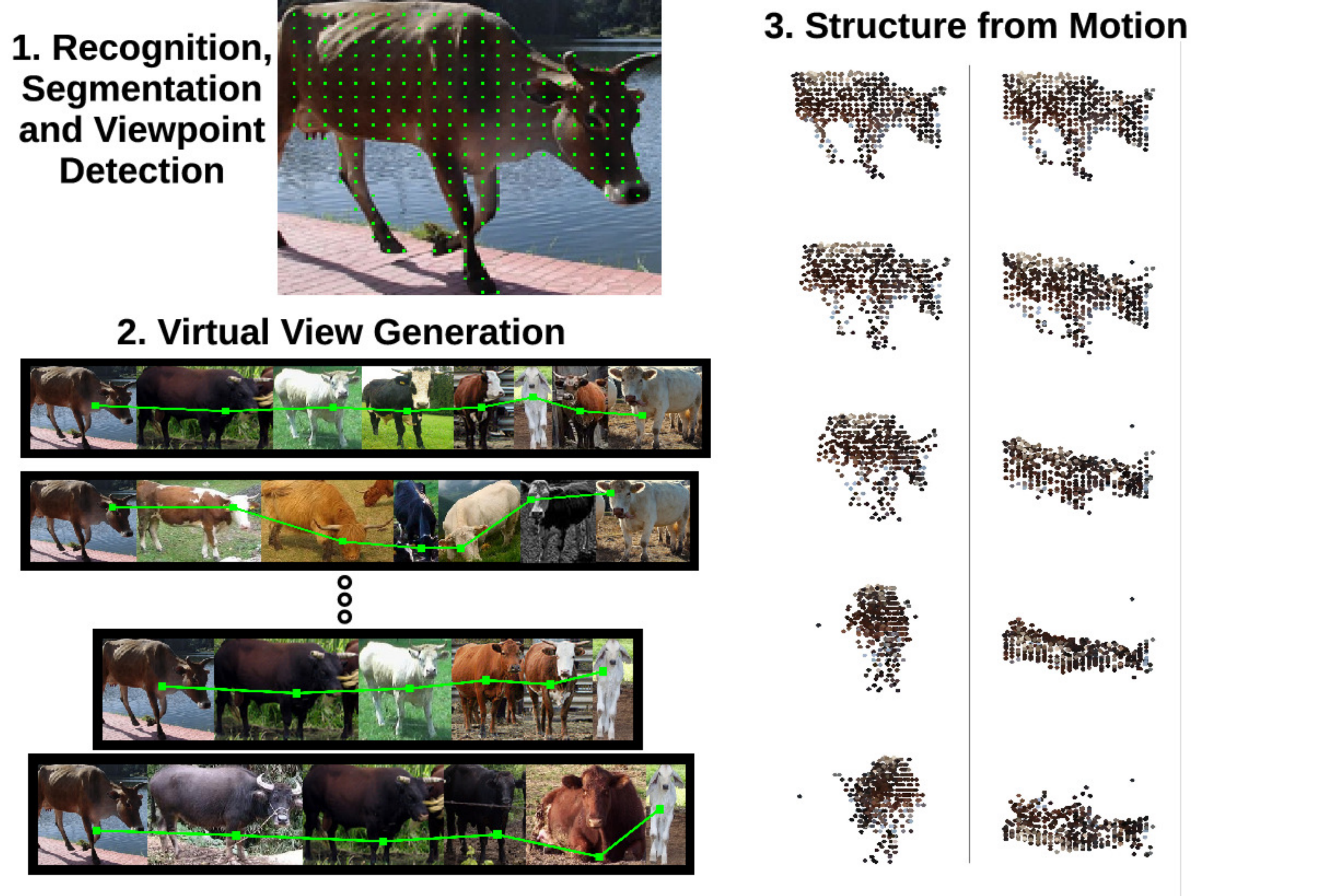}
\caption{\label{fig1} Our goal is to reconstruct an object from a single image using structure from motion techniques on virtual views obtained by aligning points on a regular grid on the test object (shown on top) with points on similar grids defined on objects in a reusable collection. Accurate alignment is achieved by computing geodesics on novel \textit{virtual view networks}, VVN in short, which induce smooth rotations through the class object collection and simplify matching. Our approach assumes object classification, localization and viewpoint detection as inputs and produces a point cloud (here shown for different camera azimuths on the left and different elevations on the right). Better seen on a computer screen with color and zoom.}
\end{figure}

Here we aim to extend SfM-style techniques to these cases by incorporating recognition. Once an object is recognized into some potentially broad class such as "cars" or "aeroplanes", one can leverage a reusable collection of images of similar objects to aid reconstruction. This is in the spirit of recent papers on face reconstruction using automatically learned morphable models \cite{kemelmacher2011face,kemelmacher2013internet} but we target generic categories and use SfM techniques. Our main insight is the following: SfM algorithms inhabit a rudimentary visual world made of 2D points in correspondence and these are all they ``see''. In this visual world, novel views can be faked much more easily than in ours, where light complicates matters. Our idea, illustrated in fig. \ref{fig1} is  to \textit{synthesize} virtual (SfM) views of the target object by aligning it with images of  different instances from the same class then employing robust rigid SfM techniques to reconstruct its visible surfaces. This idea is compatible with findings that human perception of structure from motion is robust to small shape deformations of the object \cite{jansson1973visual} and prefers to interpret them as manifestations of a rigid object with slightly altered shape instead of a non-rigid object \cite{ullman1983maximizing}.  

The main technical challenge we face is the need to align the target object with every different object in a collection, which may be pictured with arbitrary viewpoint displacements, all the way up to 180 degrees from the viewpoint of the target object. There is no dense 2D alignment technique that we know of that is prepared for such large viewpoint variation, so we propose a new one: instead of attempting to match the target object with each object in the collection individually, we predict the pose of the target object and identify a subset of objects from the collection with similar poses -- the intuition is that these will be easier to align with. Afterwards we propagate the correspondences to all other collection objects along geodesics on our new \textit{virtual view networks}.  

An additional difficulty for structure from such virtual motions is that the tight rigidity assumptions made by epipolar geometry and fundamental matrix estimation in standard RANSAC-based SfM approaches are unlikely to hold in our setting because the objects do not have exactly the same shape. Non-rigid structure from motion \cite{bregler2000recovering,paladini2009factorization} approaches, developed for reconstruction from video, have not yet been demonstrated on deformations arising from intra-class variation for generic classes. We pursue instead scaled-orthographic factorization techniques \cite{tomasi1992shape} that are more regularized, because they have fewer parameters to optimize, and we introduce methodology for a) increasing robustness to the multitude of noise sources we have by extrapolating synthetic inliers using domain knowledge and b) increasing the specificity of the resulting reconstructions, by emphasizing information in images that we are more confident about being related to the target object. 

We will review related work in the next section, before explaining in sec. \ref{sec:networks} how we build and use virtual view networks to synthesize large sets of new views from one or more images of a target object to feed to SfM. Sec. \ref{sec:reconstruction} introduces novel techniques to robustly perform SfM from noisy virtual views and sec. \ref{sec:experiments} has results on alignment and reconstruction before the paper concludes in sec. \ref{sec:conclusion}. Source code to reproduce all results will be made available online\footnote{Videos with all our reconstructions will also be made available online. A selection can be accessed at \url{http://youtu.be/JfDJji5sYXE}.}.

\section{Related Work}
\label{sec:related_work} 

Several recent papers have exploited class-specific knowledge to improve SfM. The goal in one line of work is to create denser, higher-quality reconstructions \cite{bao2013dense,dame2013dense,hane2014class}, not to regularize SfM from few images and typically requires 3D training data. Closer to our work, Bao and Savarese proposed to reason jointly over object detections and point correspondences \cite{bao2011semantic} to better constrain SfM when there are few scene points shared by different images. Our approach differs in that it focuses on reconstructing the shape of individual objects and can reconstruct from a single image of the target object. 

Our work is also related to 2D alignment approaches\footnote{There are also several papers studying alignment using 3D models \cite{su09,pepik2012teaching,boddeti2013correlation,KrauseStarkDengFei-Fei_3DRR2013}.}, that can be divided into two camps, class-specific sparse ones, that try to localize the keypoints available in a training set \cite{cootes1995active,belhumeur2011localizing,martins2012discriminative,yang2013articulated,hejrati2014analysis}, and class-agnostic dense ones such as SIFTflow \cite{liu2011sift,rubinstein2013unsupervised} and related techniques \cite{kim2013deformable} that attempt to align directly any desired pair of images. Our alignment method sits in a middle ground as it uses class information but aligns a uniform grid of points inside each object that is much denser (several hundreds of points in practice) than the typical sets of training keypoints comprising 10 to 20 points per class. 

Approaches building networks of objects have  gained popularity in vision in the last few years \cite{malisiewicz2009beyond} and have been recently proposed for 3D reconstruction from a single image \cite{su2014estimating} but using a collection of 3D CAD models, whereas we use annotated images. Other approaches requiring some form of 3D training data have been proposed for generic \cite{barron2012shape,karsch2013boundary} and class-specific \cite{hassner2013single,cashman2013shape} object and scene reconstruction \cite{hoiem2007recovering,saxena2009make3d,eigen2014depth,ladicky2014pulling,Karsch:TPAMI:14}.

\section{Virtual View Networks}
\label{sec:networks} 

As in popular class-specific sparse alignment setups \cite{cootes1995active,belhumeur2011localizing,yang2013articulated,hejrati2014analysis}, we assume that a collection of training images $\{\textbf{I}_1, ..., \textbf{I}_N\}$ is available for each class, together with a small set of $Z$ consistent keypoints $L_i = \{\textbf{m}^i_1, ..., \textbf{m}^i_Z\}$ for each image $i$, where some of them may be missing due to occlusion. We bootstrap scaled orthographic cameras, represented by rotation matrices $\{R_1, ..., R_N\}$ from the keypoints for all images using the method from Vicente \etal \cite{vicente2014reconstructing} \footnote{Cameras for non-rigid classes are computed from a representative subset of keypoints in a rigid part such as the torso, in animal classes.}. We also assume for simplicity that all objects in a collection are segmented and that at test time the localization problem has been solved and we have a segmentation of the test object, which could be obtained using a state-of-the-art semantic segmentation algorithm \cite{carreira2012semantic,hariharan2014simultaneous} or cosegmentation \cite{rother2006cosegmentation} if multiple test images are available\footnote{This is a stronger assumption but it allows us to focus entirely on reconstruction without having to dabble with the intrincacies of segmentation. In the long run the two problems are likely to be best handled in conjunction.}. We use a fixed stride for feature extraction, resulting in a regular grid of 2D locations for matching in each image $X_i=\{\textbf{x}^i_1,..., \textbf{x}^i_M\}$ with $M$ associated descriptors $\{\textbf{d}^i_1, ..., \textbf{d}^i_M\}$, inside each object segmentation. The particular descriptors used will be described in the experimental section.

\begin{figure}[t]
\includegraphics[width=1\linewidth]{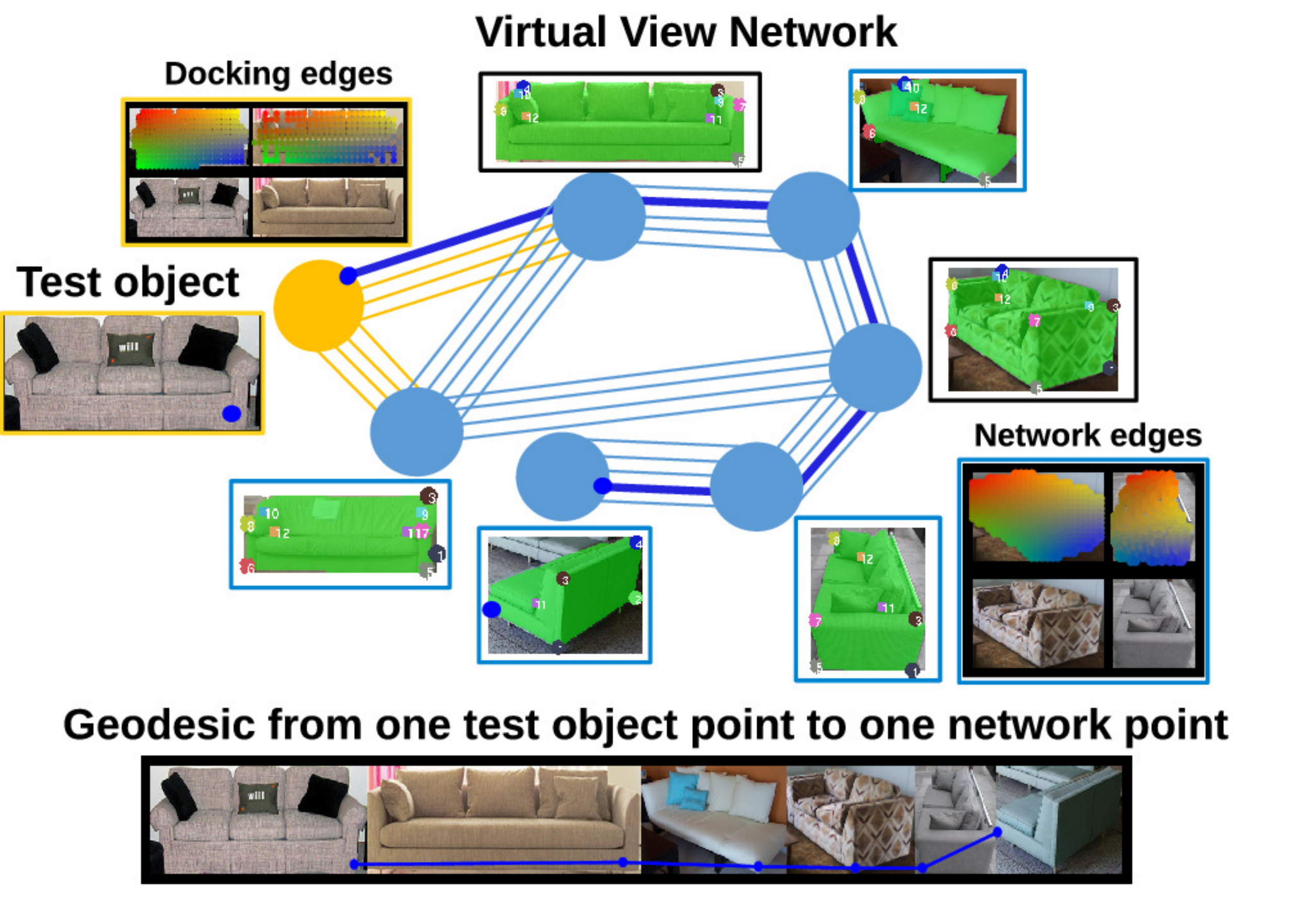}
\caption{\label{fig:network} Instead of matching a test object to each training object directly, which may be difficult due to viewpoint variation, we match through a network connecting training objects with similar viewpoint. A test object is docked to the network by matching it to a few network objects with similar viewpoint ($10$ in practice), then it is aligned with all other objects based on geodesic distances in the network. Points connected by an edge are shown with similar color.}
\end{figure}

SfM algorithms operate on a set of point tracks, which were traditionally obtained by tracking local features in video frames and later also from unstructured image collections \cite{snavely2006photo}. Here we aim to compute one track for each feature in the target object by matching it to corresponding features in "virtual views" borrowed from every object in the training collection, a hard problem because local appearance changes dramatically with viewpoint variation. We convert this hard wide-baseline problem into many easier small-baseline ones by defining a distance between feature points that considers a network over the whole collection of objects. Let network $G=\{\mathcal{V},\mathcal{E}\}$ have nodes for all points in $\{X_i, ... X_N\}$ and edges derived by matching points in objects having similar pose. We dock the image of the test object to the network by matching it to a few objects also chosen based on pose, which is assumed to be computed using a pose detector for the test object, and compute geodesics (shortest paths) between each point in the target object and all points in the collection, which can be done efficiently \cite{fredman1987fibonacci} using Dijkstra's algorithm $M$ times, one for each feature in the target object. This network distance can then be used as a more meaningful alternative to standard euclidean distance based on appearance features, for matching the test object to all training objects (using for instance nearest neighbor matching). 
The overall idea is illustrated in fig. \ref{fig:network}.

\subsection{Network Construction}

We match separately each object in the collection to a fixed number of nearest neighbors in pose space ($30$ in practice), measured using the riemannian metric on the manifold of rotation matrices $||log(R_i R_j^T)||_F$, where $log$ refers to the matrix logarithm and $||.||_F$ denotes the Frobenius norm of the matrix.
 \textit{Drifting} is a major concern in any tracking approach and is especially hard to deal with automatically in our case, over different objects. We counter drifting by regularizing feature matching using symmetric warping priors derived from the manually annotated keypoints. Let $\alpha$ be a weighting parameter. We define the cost of matching points $u$ and $v$ with locations $\textbf{x}^i_u$ and $\textbf{x}^j_v$ with descriptors $\textbf{d}^i_u$ and $\textbf{d}^j_v$ in two different images $i$ and $j$ as: 

\begin{multline}
\label{matching_cost}
E(u,v) = ||\textbf{d}^i_u - \textbf{d}^j_v|| + \alpha \cdot \big[E_w(\textbf{x}^i_u,\textbf{x}^j_v, L_i,L_j) \\ + E_w(\textbf{x}^j_v,\textbf{x}^i_u,L_j,L_i)\big].
\end{multline}

\noindent and model the warping cost using an interpolant $g:\mathbf{R}^2\rightarrow\mathbf{R}^2$, here instantiated as thin plate splines which are popular for modeling various types of shapes in vision and can be fit in closed form \cite{belongie2002shape,bookstein1997morphometric}. We define the warping cost as: 

\begin{equation}
E_w(\textbf{x}^i_u, \textbf{x}^j_v, L_i, L_j) = ||\textbf{x}^j_v - g_{ij}(\textbf{x}^i_u)||
\end{equation}

\noindent where $g_{ij}$ is fit to map $L_i$ to $L_j$. We use two warping costs for symmetry, one in each direction, as we found this to lead to more accurate alignment in practice. 

Given matching costs between all pairs of points in two neighboring objects, we add a directed edge to the network from each node $u$ to each node $v$ satisfying $\argmin_v E(u,v)$. 

\subsection{Docking to the Network} 

We do not use thin plate splines for computing matching costs when docking test instances to the network because this would require keypoints that are unavailable at test time. Cost functions similar to those used in optical flow \cite{liu2011sift} would be valid alternatives, but incur some computational cost. We opted instead to simply replace the thin plate splines in eq. \ref{matching_cost} by affine interpolants $h:\mathbf{R}^2\rightarrow\mathbf{R}^2$ fit to map between the $4$ corners of the bounding boxes of the target object and a docking object, resulting in an anisotropic scaling. This makes sense because it biases corresponding points to be in the same relative location within a bounding box, a good prior since we are docking objects with similar viewpoint. Note that a single spatial term is sufficient in this case, because the mapping is symmetric. Matching proceeds as when constructing the network, but we suppress multiple edges connecting to the same point in a docking object and keep only the one with minimum weight, to enable the speed-up to be presented next. 

\subsection{Fast Alignment at Test Time}

It is usually desirable to push as much as possible the burden of computation to an offline stage and to have fast performance at test time. This is also feasible with our method, assuming nearest neighbor matching is used and hence we only need to identify points in training objects having minimum geodesics to points in test objects (e.g. retrieving distances to all points in each training object is unnecessary) by using the fact shortest paths are recursive. We precompute the nearest neighbor matchings using network distances between all pairs of objects in the network then construct a new network with the same nodes, but with edges directly set between all pairs of matching points, with weights being the geodesic distances in the original network. On the new network all shortest paths between points in any two objects can be identified by simply selecting the outgoing edges having minimum weight, a property we will call being point-to-point.

At test time one must consider an additional set of edges between points in the test object and points in docking objects and the network becomes no longer point-to-point. Assuming there is at most a single edge from a test object point to each network node in docking objects, however, this edge can be pushed forward and summed to all outgoing edges from nodes in docking objects making the network again point-to-point and geodesics from each test point to all points in all objects can be found by selecting the minimum edges from any of its docking points, a linear time operation in the number of network nodes, which compares favorably to Dijkstra's quadratic time. Using this technique we manage to align a test object to a collection of roughly $1000$ objects having $300,000$ points in around half a second on a modern desktop computer, instead of in more than a minute using Dijkstra's algorithm. 


\section{Reconstruction}
\label{sec:reconstruction} 

Reconstruction faces three challenges in our setting: integrating sparse, far apart views of the target object, coping with noise in virtual views synthesized from training objects and producing a shape that is specific to the target object while pooling shape evidence from all objects in the training collection. 

We use all network images to reconstruct each test instance, and deal with noise by assuming that all generated virtual views are of a same rigid object undergoing rigid motion under scaled-orthographic projection, which has the positive effect of allowing us to estimate fewer parameters than in non-rigid reconstruction or formulations assuming perspective projection and to adopt the well-studied Tomasi-Kanade factorization framework \cite{tomasi1992shape}. More specifically we employ the very robust Marques and Costeira algorithm \cite{marques2009estimating} which can handle missing data. 

Sparse reconstruction from many views of an object is an almost solved problem \cite{hartley2003multiple,snavely2006photo,furukawa2010accurate}. Here we focus on reconstruction from few views of the target object, in particular from a single image plus its mirrored version, which, because most object classes (e.g. cars, aeroplanes) possess bilateral symmetry, provides a second image from a different viewpoint for free. Directly matching the original and mirror view is however not feasible in general, for example a side view of a car from the right has almost no points shared with a side view from the left side. We propose \textit{network-based factorization} to handle these issues. To cope with outliers we introduce a technique called \textit{synthetic inlier extrapolation} and, finally, we propose two strategies for building up specificity in the reconstruction towards the shape of the target object, \textit{resampling} and \textit{xy-snapping}. We will begin by describing synthetic inlier extrapolation.

\subsection{Synthetic Inlier Extrapolation}
\label{sec:inlier}

Even though factorization has few parameters compared to approaches based on bundle adjustment, they can still be negatively affected by outliers. There is prior work on handling known gaussian noise distributions \cite{aguiar2003rank,irani2000factorization} and outliers \cite{xu1995robust, de2003framework} within factorization, but these approaches may not be trivial to adapt so they can deal with missing data. Here we propose instead to reduce the influence of outliers by \textit{swamping} the data with synthetic inliers generated using domain knowledge, namely we sample a constant number ($10$ in practice) of equally spaced points along 2D lines connecting all pairs of ground truth keypoints in the training images. Such points define correct correspondences between different images (as much as they can, ignoring object shape variation) under scaled orthographic projection. To see this, let $\textbf{p}_u,\textbf{p}_v$ be the coordinates of two keypoints in the 3D shape with $\textbf{m}_u = M \textbf{p}_u, \textbf{m}_v = M \textbf{p}_v$ being  the corresponding image coordinates, where $M$ is the projection matrix. Let $\textbf{p}_{uv}(\alpha)$ = $\alpha  \textbf{p}_u + (1-\alpha) \textbf{p}_v$ denote a point on the line connecting the two keypoints. It can be shown that under orthographic assumptions, $M \textbf{p}_{uv}(\alpha) =  \alpha \textbf{m}_u + (1-\alpha) \textbf{m}_v$. 
  
\begin{figure}
\centering
\renewcommand{\arraystretch}{1}
\begin{tabular}{@{}c@{} c@{}}
\includegraphics[width=0.45\linewidth]{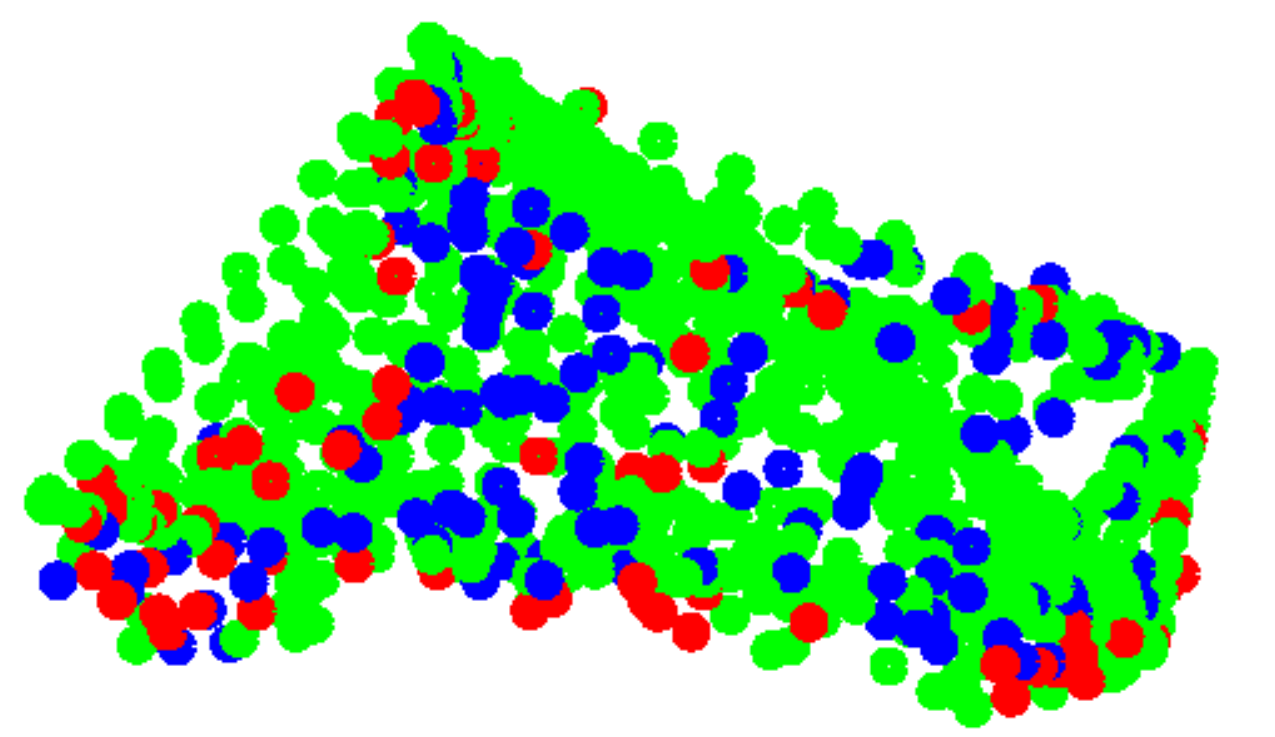} &
\includegraphics[width=0.45\linewidth]{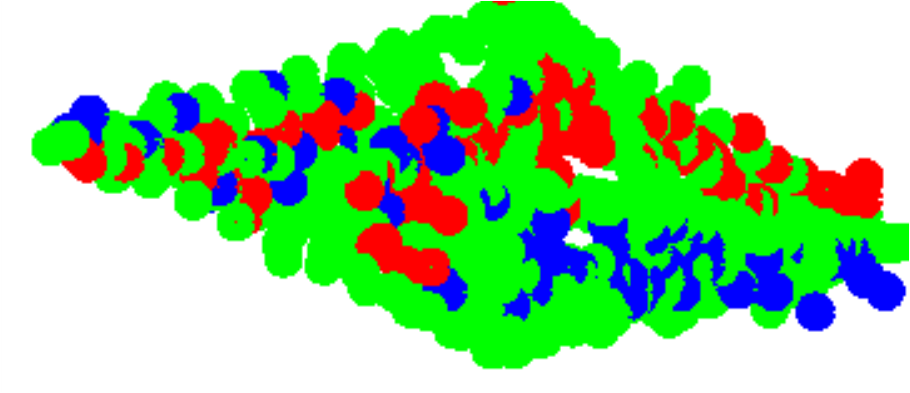} \\
\includegraphics[width=0.45\linewidth]{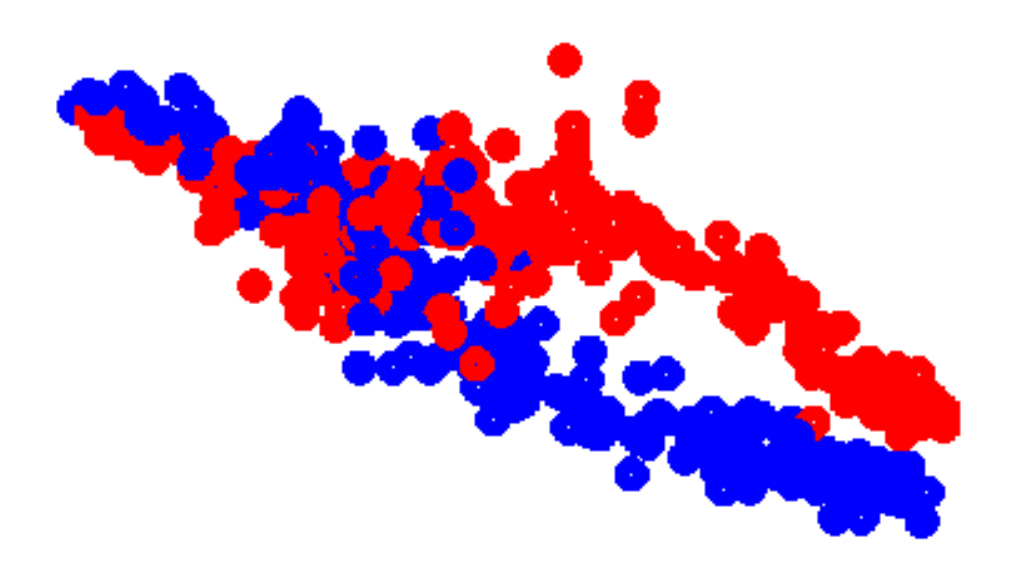} &
\includegraphics[width=0.45\linewidth]{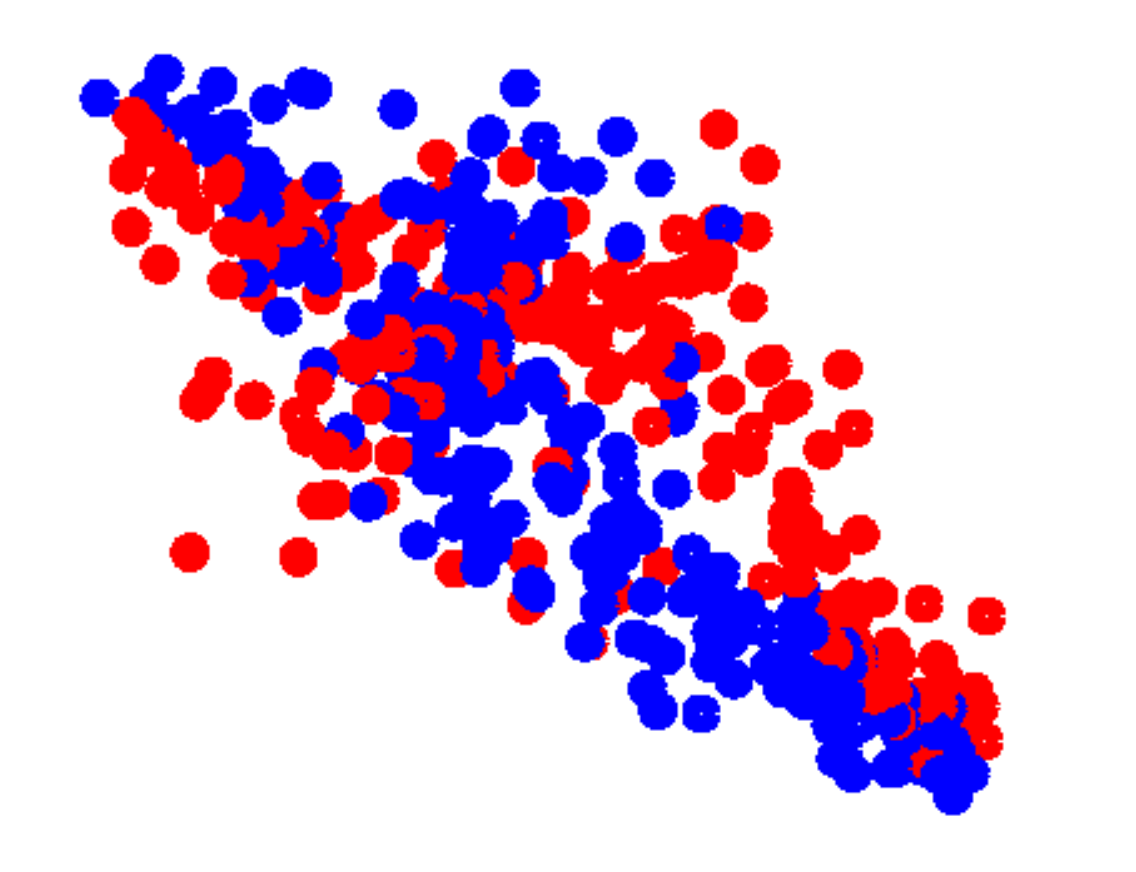} \\
\end{tabular}
\caption{\label{fig:inlier} The effect of reconstructing points from an image (blue) and its mirrored version (red) with and without extrapolated synthetic inliers (in green). Shown on the top row are all reconstructed points for a motorbike using extrapolated inliers, seen from side and top, and on the bottom row are shown top views of the computed shapes using extrapolated inliers (left) and not using extrapolated inliers (right). The motorbike becomes much wider and noisier without extrapolated inliers.}
\end{figure}

\subsection{Network-Centered Factorization}

Classic rigid factorization builds an observation matrix having two rows for each of $N$ frames in an input video sequence and one column for each of $K$ tracked points:  
\begin{equation}
W = 
\begin{bmatrix}
x^1_1& \cdots & x^1_K\\
y^1_1& \cdots & y^1_K\\
\vdots \\
x^N_1& \cdots & x^N_K\\
y^N_1& \cdots & y^N_K\\
\end{bmatrix},
\end{equation}

\noindent then compute a $3\times K$ shape $S$ as well as rotation matrices, translation vectors, and scale parameter for each image from this matrix. 

\begin{figure}[t]
\includegraphics[width=0.8\linewidth]{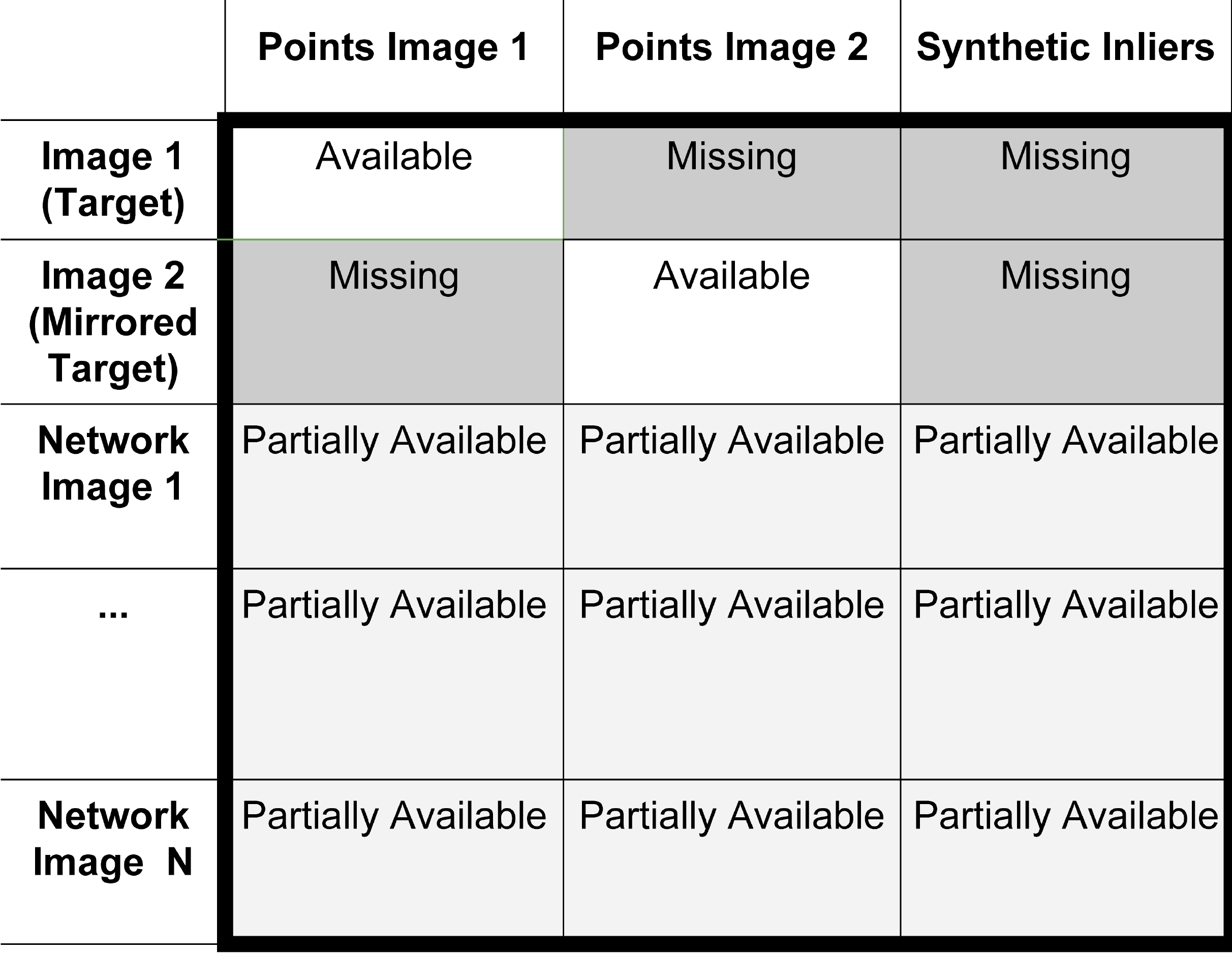}
\caption{\label{fig:missing_data} The blockwise pattern of missing data in the observation matrix (inside the bold lines) for our network-centered factorization approach, here instantiated in the case where two images of the object are docked to the network -- if more images are available they can be used in the same way. The SfM algorithm fills in the missing data so we retrieve the reconstructed points for all images from the first block-row shown in this table and ignore the reconstructed points of synthetic inliers, which are only used as regularization.}
\end{figure}

In our case, each column will contain the coordinates of one point in the target object and the coordinates of those points in network objects that are aligned to it. Our observation matrix has a more specific structure as well, shown in fig. \ref{fig:missing_data}, motivated by our reliance on the virtual view network as an alignment hub which multiple target images can be docked to, hence the name network-centered factorization. We create one set of distinct points for each image of the target object, because we do not know a priori if points are shared by multiple views, and fill in tracks only between points in target images and images in the network (e.g. we do not match the target images directly), then set the rest of the matrix as missing data for the factorization algorithm to fill in. The extrapolated synthetic inliers are also added as separate points which are available for the training images but not for the test images, where they are also set as missing data. We use these points just as an additional source of regularization and ignore their reconstruction afterwards and this may be better understood by consulting fig. \ref{fig:inlier}. 

\subsection{Building up Target Specificity} 
\label{sec:specificity}
We use two strategies for increasing the specificity of the reconstruction towards the target object: resampling and xy-snapping. 

\vspace{2mm}
\noindent \textbf{Resampling.} Factorization algorithms compute low-rank matrix approximations and these can be weighted so that some of the observations are given more importance. Sophisticated algorithms \cite{srebro2003weighted} for this task have been developed but not yet demonstrated on structure from motion. Here we propose instead to boost the importance of the target images and a few nearest neighbors from the training set (based on appearance), by simply resampling their rows in the observation matrix. This is equivalent to finding the low-rank factorization that minimizes a weighted euclidean loss with the rows corresponding to the important instances having higher weights.

\vspace{2mm}
\noindent \textbf{xy-Snapping.} The points from the target object are the only ones we can trust blindly and which should be considered correct. We enforce this as post-processing by snapping the points of the target object back to their original coordinates in a reference image after reconstruction. If we are reconstructing also using a mirrored image, we can compute correspondences across the symmetry axis trivially by tracking where points move to during mirroring, and then just translating them in the image plane by the same offset as the points in the original image. 


\section{Experiments}
\label{sec:experiments} 

Our focus is on alignment and reconstruction so we will assume that target objects have been localized and segmented as discussed in the introduction. We will evaluate 2D alignment and reconstruction separately, in each of the following subsections. We will study the impact of the accuracy of pose prediction on 2D alignment and will assume viewpoint has been correctly detected in the reconstruction section. All experiments used PASCAL VOC \cite{everingham2010pascal}, where there are segmentations and around $10$ keypoints available for all objects in each class \cite{hariharan2011semantic}. The same setup and the same $9,087$ fully visible objects were used as in \cite{vicente2014reconstructing}, but we split them into $80\%$ training data and $20\%$ test data and built virtual view networks on the training images and their mirrored versions, and evaluated alignment performance on test data without using keypoints. We discarded classes "dining table", "bottle" and "potted plant" in the reconstruction section because their keypoints are marked in a view-dependent way (e.g. bottles have keypoints marked on their silhouettes, so the induced cameras are always frontal and direct depth recovery requires additional cues). 

\subsection{2D Alignment}

We resized the image of each object to be $150$ pixels tall and computed a regular grid of features taken by concatenating the fourth and fifth convolutional layers of the AlexNet convolutional network \cite{krizhevsky2012imagenet}, resulting in 640 dimensional feature vectors at each grid location. We obtained a stride of 8 pixels by offsetting the image appropriately and passing it through AlexNet multiple times, then carefully assembling back the multiple resulting grids (similar to \cite{sermanet2013overfeat}). We also evaluated SIFT features, computed with a stride of 2 pixels and all feature extraction was performed with the background pixels set to black \footnote{We convert images to be gray-valued and compress the pixel value range to be between 30 and 255 before zeroing out the object background, in order to preserve contrast along the boundaries of dark objects}. Each object's figure-ground segmentation was also used for ignoring grid points in the background during matching.

Our full proposed approach, VVN, aligns a test object to each training object using nearest neighbor matching on a distance function defined by geodesics on a network connecting all grid points on all training examples. While there are many class-specific techniques for localizing a set of keypoints available in training data, we are not aware of techniques of that kind that are able to align arbitrary grids of points. We opted then to compare with techniques that can align grids even though they do not use class-specific knowledge: nearest neighbor matching using the euclidean distance and SIFTflow \cite{liu2011sift}, using either SIFT or the same deep features we employ. We evaluate alignment by matching each test image to all training images and checking how the ground truth test keypoints match to the training image keypoints. We average the following per-pair matching error:

\begin{equation}
\label{loss}
L(C) = \frac{1}{B} \sum_u ||\textbf{c}_u - \textbf{m}_{u}||, 
\end{equation}

\noindent where $u$ iterates over the grid points closest to each ground truth keypoint $u$ on the test image, $C$ is the set of corresponding points ${c_1,...,c_B}$ on a training image according to the matching, and $\textbf{m}_{u}$ is the position of ground truth keypoint $u$ on the training image. We average the errors over all images in all $20$ PASCAL classes.

\begin{figure}[t]
\centering
\includegraphics[width=0.95\linewidth]{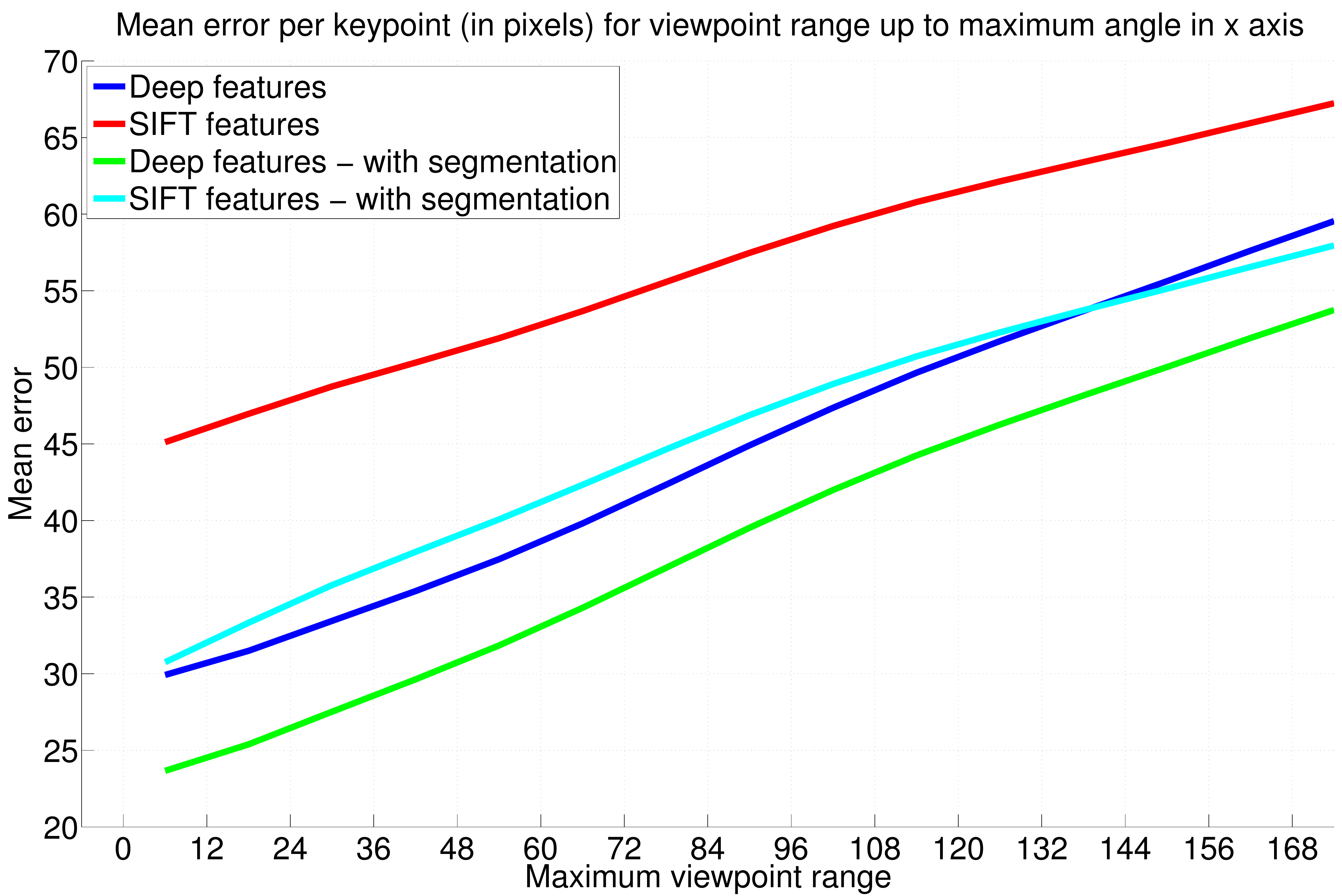}
\caption{\label{fig:mask_vs_nomask} Mean error in eq. \ref{loss} when matching points in two objects from the same class using nearest neighbor, as a function of the viewpoint difference between the objects. Deep features allow for more accurate alignment, and this is more evident when segmentation is not available.}
\end{figure}

\vspace{2mm}
\noindent \textbf{Different Features and Segmentation.} We first compared SIFT and deep feature matching using nearest neighbor with the euclidean distance, and evaluated how important it is to have segmentation for this task. Deep features do better in general and the matching errors are slightly worse without segmentation, especially when matching with SIFT features - the deep features seem better prepared in the presence of background clutter which is promising but we will assume segmentation for the rest of the paper. These results are shown in fig. \ref{fig:mask_vs_nomask}. 

\begin{figure}[t]
\centering
\includegraphics[width=0.95\linewidth]{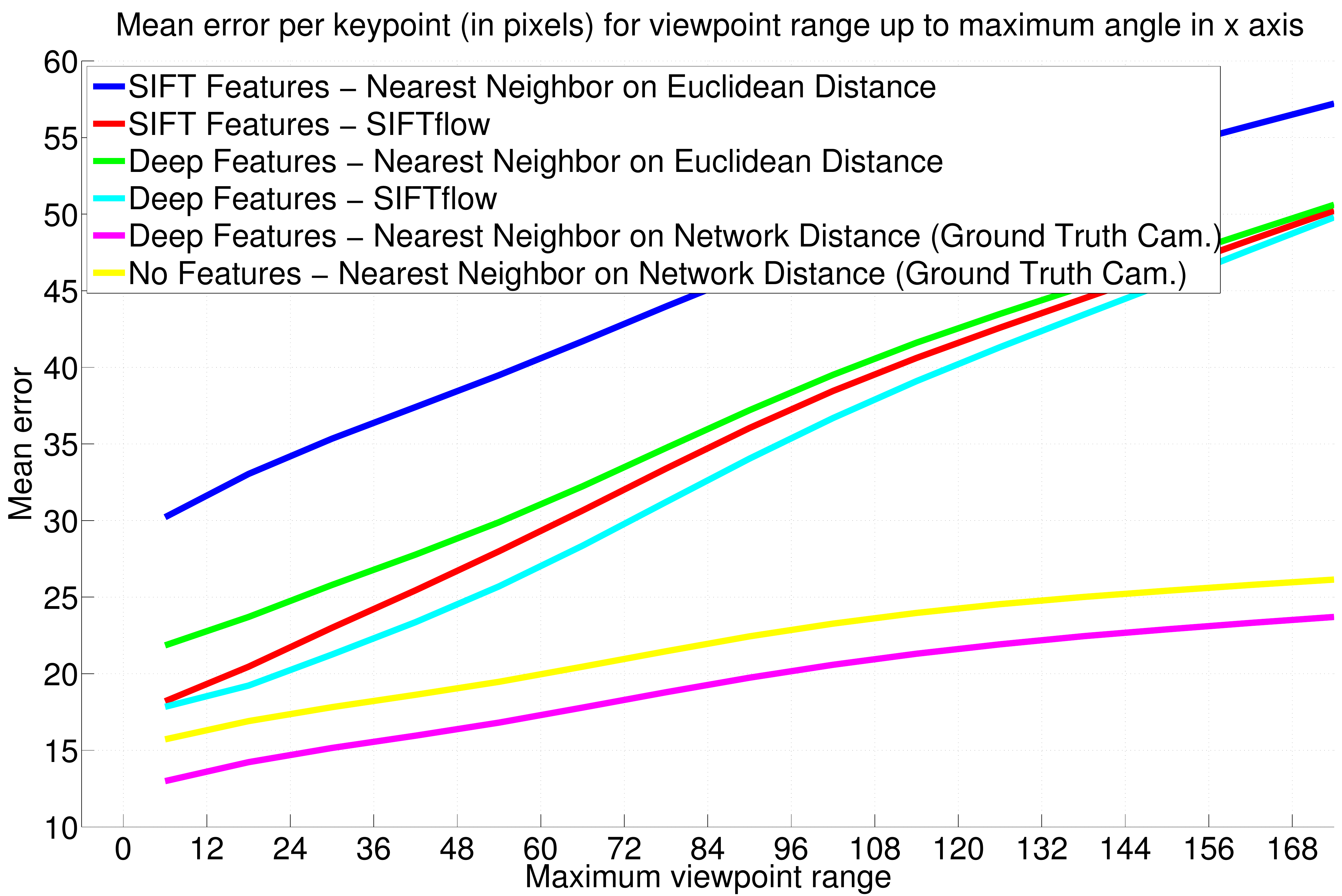}
\caption{\label{fig:siftflow_nn} Mean error when matching points in two segmented objects from the same class using nearest neighbor with euclidean distance and SIFTflow, compared to nearest neighbor and our proposed network distance, as a function of the viewpoint difference between the objects. SIFTflow improves considerably over nearest neighbor matching using euclidean distance but is not robust to large viewpoint variation. Results are good even using a network built without features, using just the spatial terms.}
\end{figure}

\vspace{2mm}
\noindent \textbf{Euclidean and Network Distance.} We compare our proposed nearest neighbor matching using the network distance to two baselines, nearest neighbor with the euclidean distance and SIFTflow. All methods used images with the background masked and the correspondences constrained to be inside the segmentation. In this experiment we assume knowledge of ground truth cameras for selecting the elements in the network to dock the test object with, which we will also assume in the reconstruction section. The results are shown in fig. \ref{fig:siftflow_nn} and demonstrate that given an accurate pose estimate for the test object, the network distance leads to more accurate alignment up to the maximum 180 degrees viewpoint difference. SIFTflow leads to large gains over nearest neighbor using SIFT features and euclidean distance but is less robust to viewpoint variation.

\begin{figure}[t]
\centering
\includegraphics[width=0.95\linewidth]{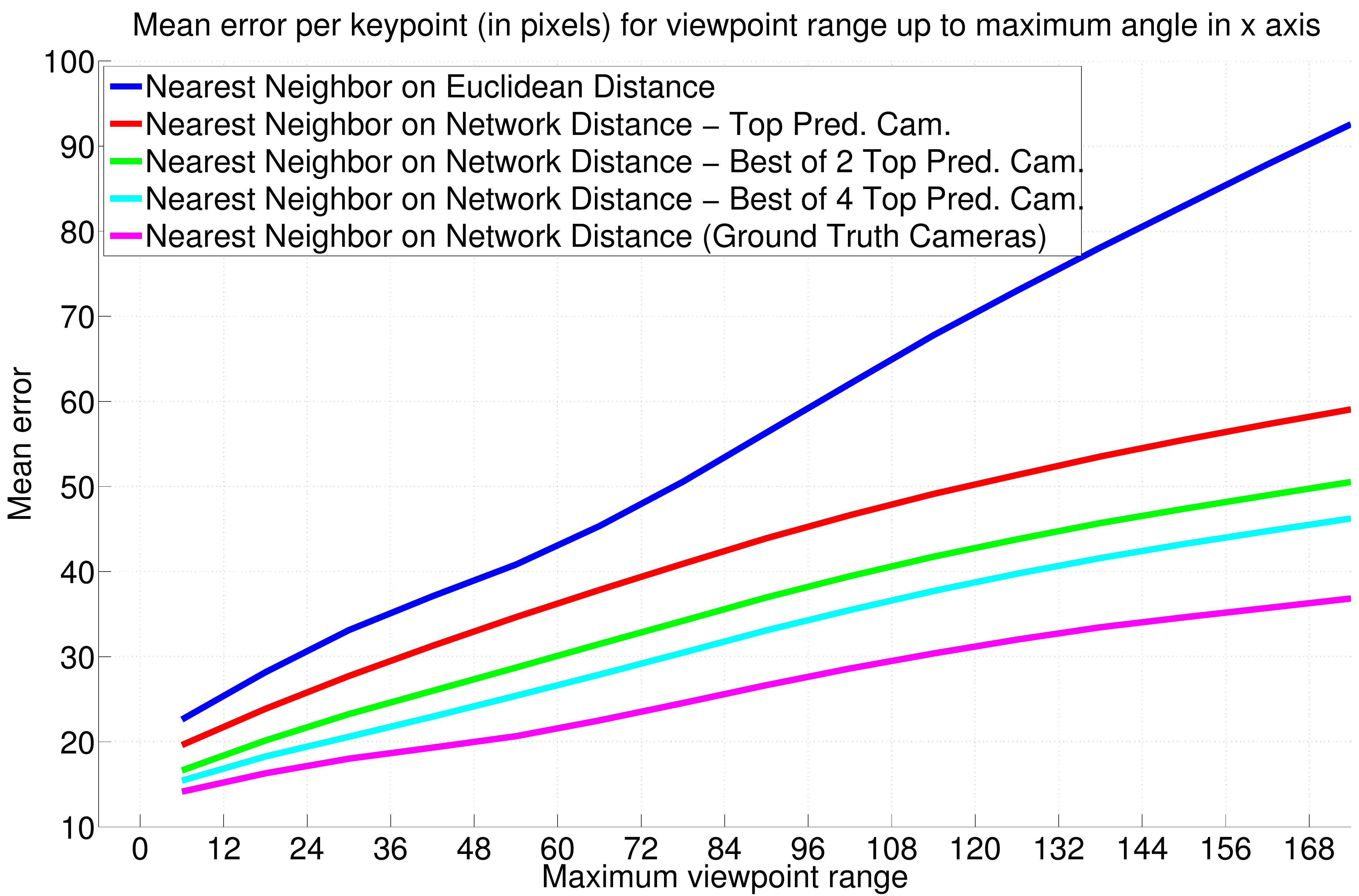}
\caption{\label{fig:network_vs_direct} Mean error when matching points in two segmented objects from the same class using nearest neighbor with euclidean distance and our proposed network distance, as a function of the viewpoint difference between the objects. The network distance leads to more accurate alignment across all viewpoint differences. The same deep features are used in all cases here and we consider only rigid categories, for which there is pose prediction training data from Imagenet. See text for details.}
\end{figure}

\vspace{2mm}
\noindent \textbf{Pose Prediction and Alignment.} Our final and main experiment in this subsection evaluates VVN alignment when using automatic pose prediction. We discretized the space of rotations into 24 equally spaced bins and learned a convolutional network classifier jointly over all classes, using the AlexNet architecture with original weights finetuned using the pose annotations from PASCAL3D+ \cite{xiang2014beyond} which has many additional training examples from Imagenet for the 12 rigid categories in PASCAL VOC. The alignment results are shown, for the 12 rigid categories, in fig. \ref{fig:network_vs_direct} and demonstrate that the improvements over nearest neighbor with the euclidean distance still hold with automatic pose prediction. We also measured accuracy when using the best among the 2 and 4 top-scoring predicted poses and found this to bring large improvement, which suggests pose reranking as an interesting direction for future work. We show sample alignments for our method and siftflow on the same grid of deep features in fig. \ref{fig:alignments}.

\begin{figure}
\centering
\renewcommand{\arraystretch}{1}
\begin{tabular}{@{}c@{} c@{}}
\includegraphics[width=0.45\linewidth]{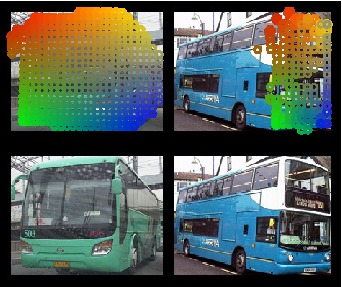} &
\includegraphics[width=0.45\linewidth]{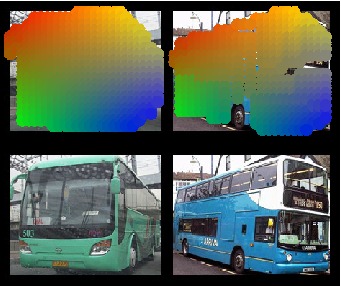} \\
\includegraphics[width=0.45\linewidth]{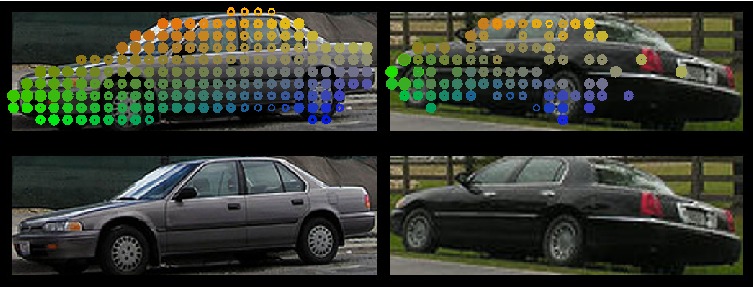} & 
\includegraphics[width=0.45\linewidth]{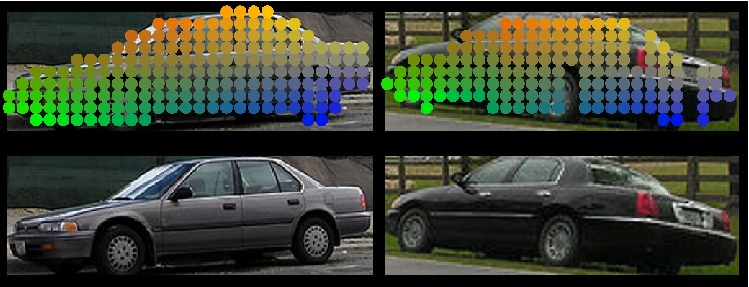} \\
\includegraphics[width=0.45\linewidth]{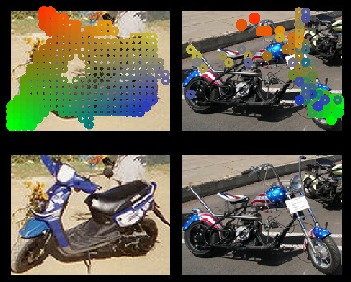} & 
\includegraphics[width=0.45\linewidth]{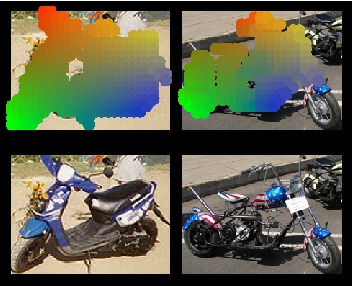} \\
\end{tabular}
\caption{\label{fig:alignments} Example alignments using our proposed network-based approach, VVN, with automatic pose prediction (first two columns) and SIFTflow (last two columns), on the same grid of deep features and assuming correct figure-ground segmentation. Corresponding points are colored the same. VVN exploits class-specific knowledge and pose prediction to obtain resillience to viewpoint variation. See the text for additional details and the supplementary material for images showing  other alignments.}
\end{figure}

\subsection{Reconstruction}

We reconstructed PASCAL VOC objects in the test set of each class, producing fuller 3D reconstructions from a single view by taking advantage of bilateral symmetry as discussed in sec. \ref{sec:reconstruction}. We used the same parameters for all classes, except xy-snapping which helped noticeably in most cases but degraded subtle aeroplane wings and bicycle handles so we disabled it on these two classes - the only class-specific option we introduced. We resampled the target image and its mirrored version $0.05$ times the number of training examples, and their $4$ nearest neighbors from the training set $0.02$ times the total number of training examples for the class (see sec. \ref{sec:specificity}). Nearest neighbors were computed from those training examples in the $30$ to $60$ degree range of viewpoint differences to the pose of the test example, selected based on euclidean distance between descriptors obtained using second-order pooling \cite{carreira2012semantic} on the AlexNet layer 5 features. The idea was to discard the spatial information in the layer 5 grid to better cope with viewpoint variation. 

\begin{figure*}
\renewcommand{\arraystretch}{1}
\begin{tabular}{@{}c@{} c@{} c@{} c@{} c@{} c@{} c@{} c@{} c@{} c@{}}
\includegraphics[width = 0.1\textwidth, keepaspectratio = true]{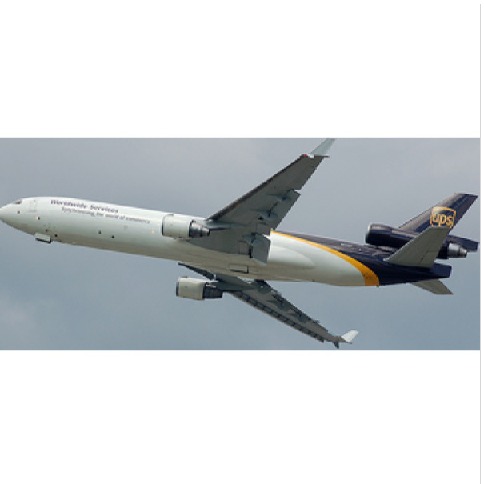} &
\includegraphics[width = 0.1\textwidth, keepaspectratio = true]{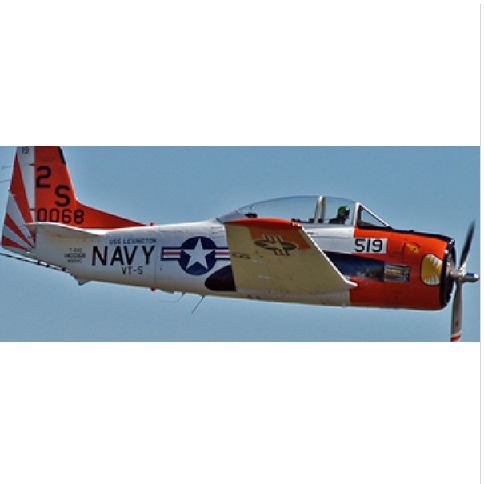} &
\includegraphics[width = 0.1\textwidth, keepaspectratio = true]{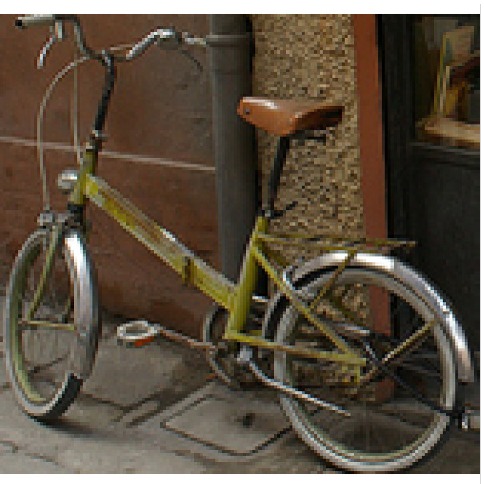} &
\includegraphics[width = 0.1\textwidth, keepaspectratio = true]{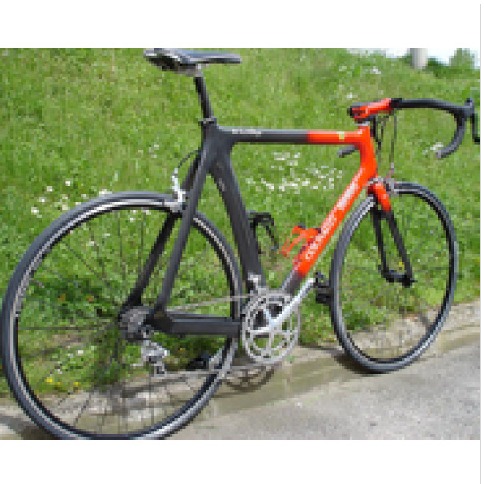} &
\includegraphics[width = 0.1\textwidth, keepaspectratio = true]{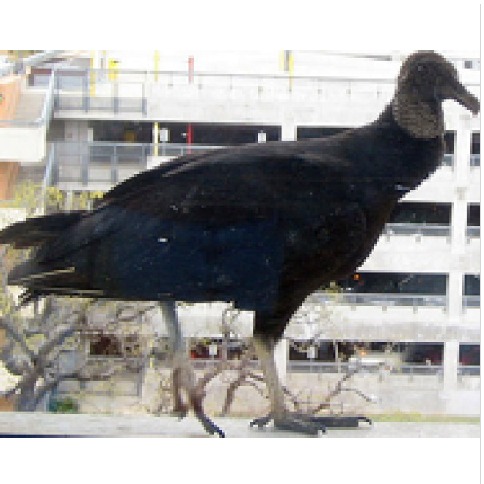} &
\includegraphics[width = 0.1\textwidth, keepaspectratio = true]{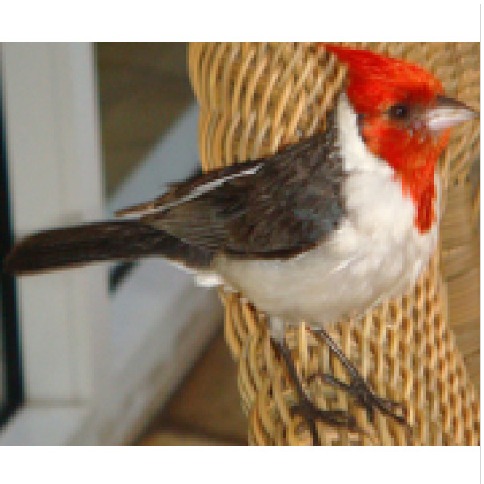} &
\includegraphics[width = 0.1\textwidth, keepaspectratio = true]{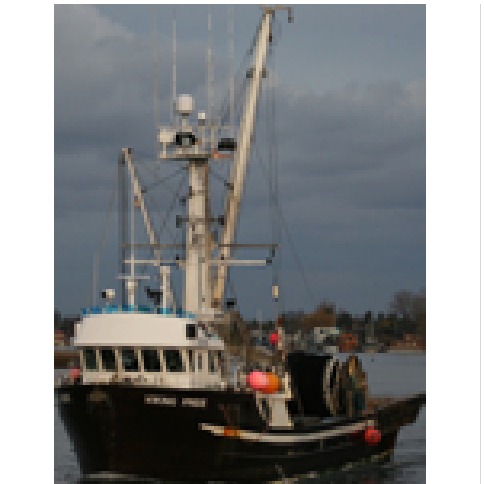} &
\includegraphics[width = 0.1\textwidth, keepaspectratio = true]{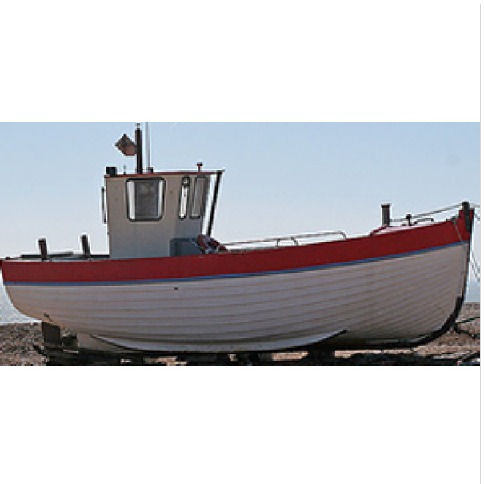} & 
\includegraphics[width = 0.1\textwidth, keepaspectratio = true]{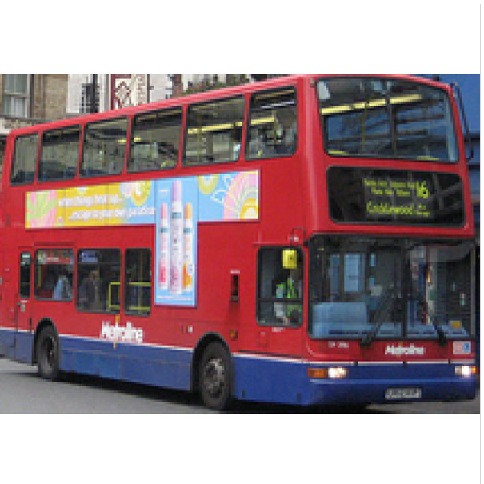} &
\includegraphics[width = 0.1\textwidth, keepaspectratio = true]{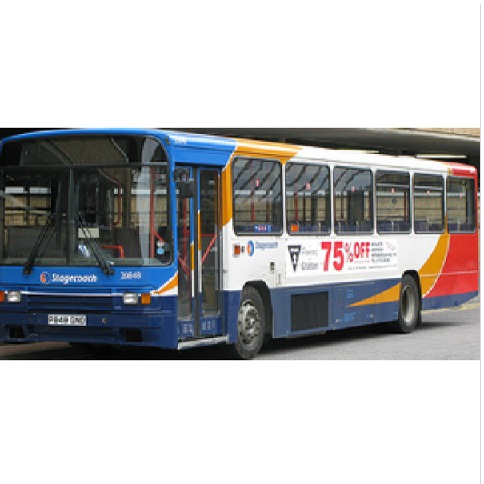} \\
\includegraphics[width = 0.1\textwidth, keepaspectratio = true]{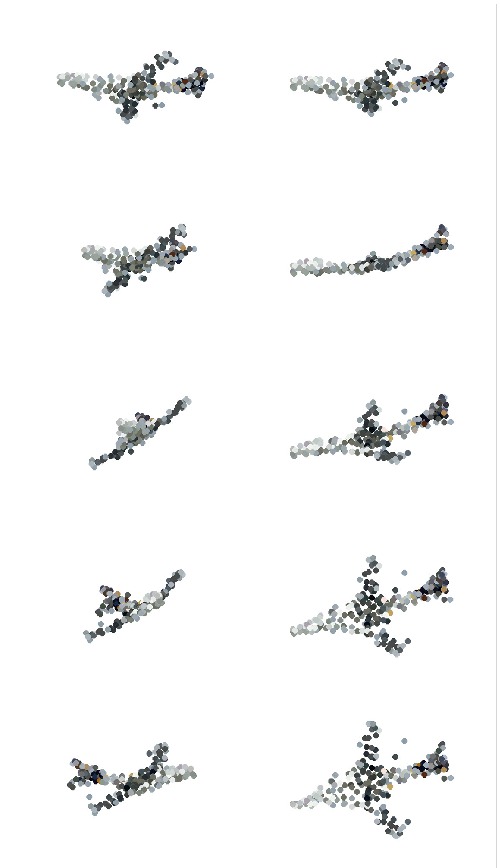} &
\includegraphics[width = 0.1\textwidth, keepaspectratio = true]{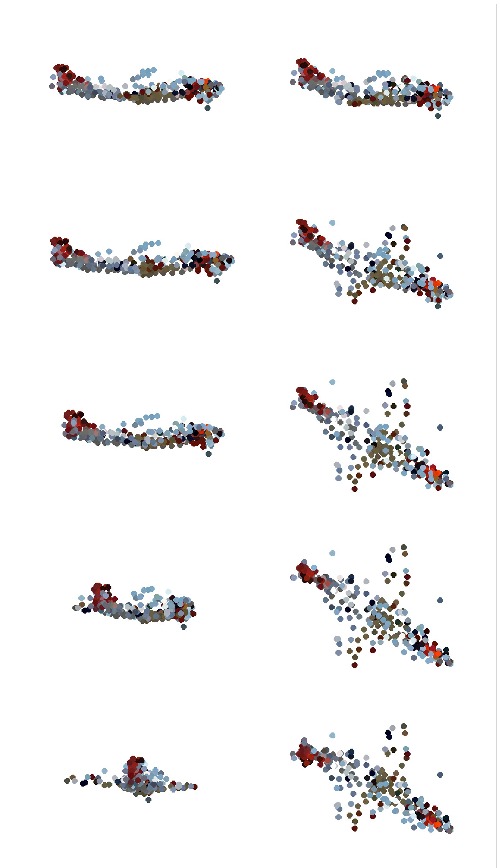} &
\includegraphics[width = 0.1\textwidth, keepaspectratio = true]{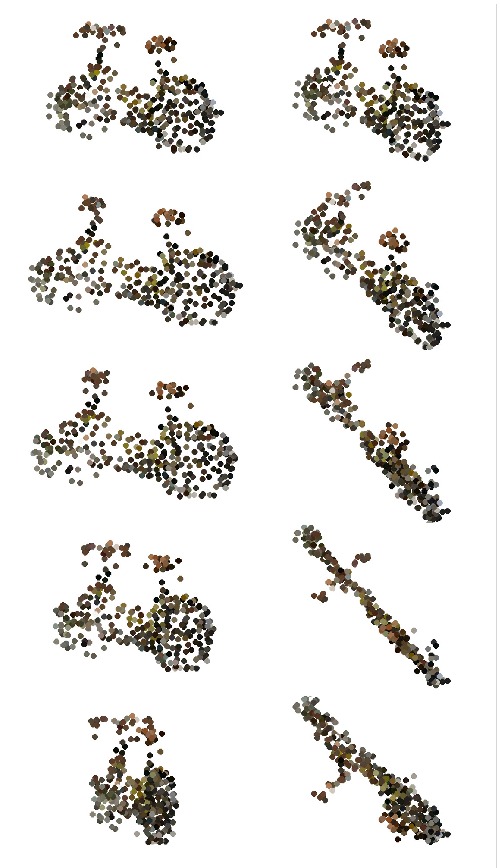} &
\includegraphics[width = 0.1\textwidth, keepaspectratio = true]{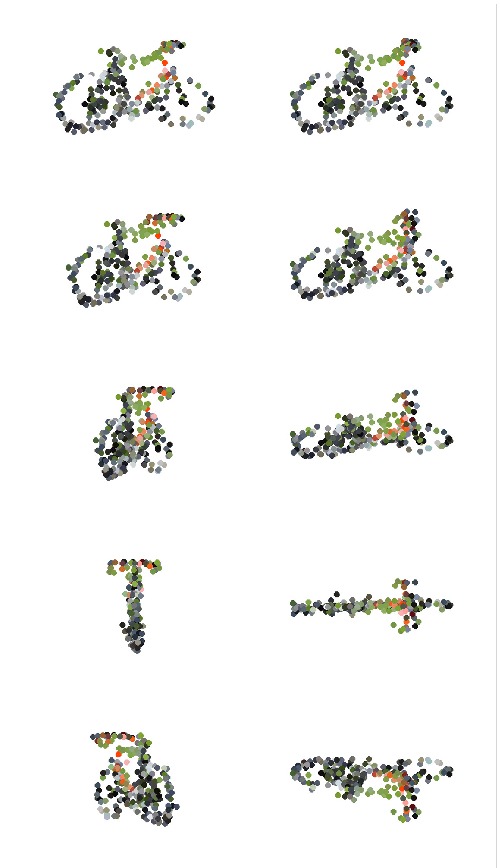} & 
\includegraphics[width = 0.1\textwidth, keepaspectratio = true]{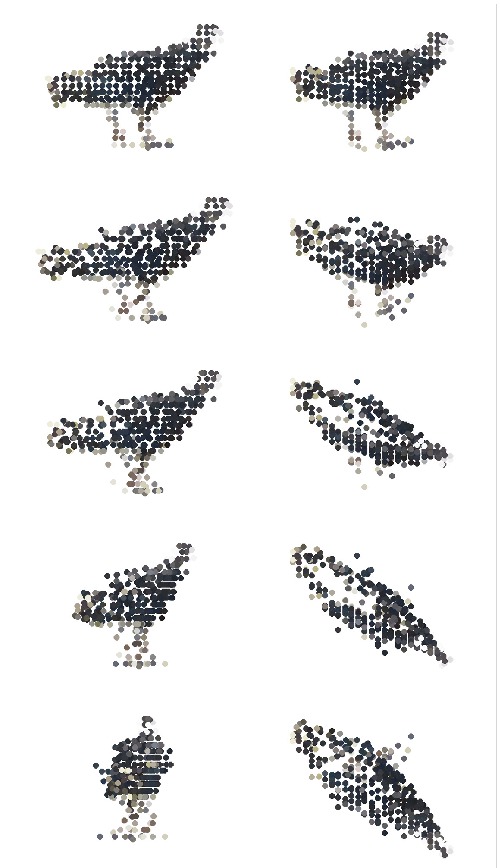} &
\includegraphics[width = 0.1\textwidth, keepaspectratio = true]{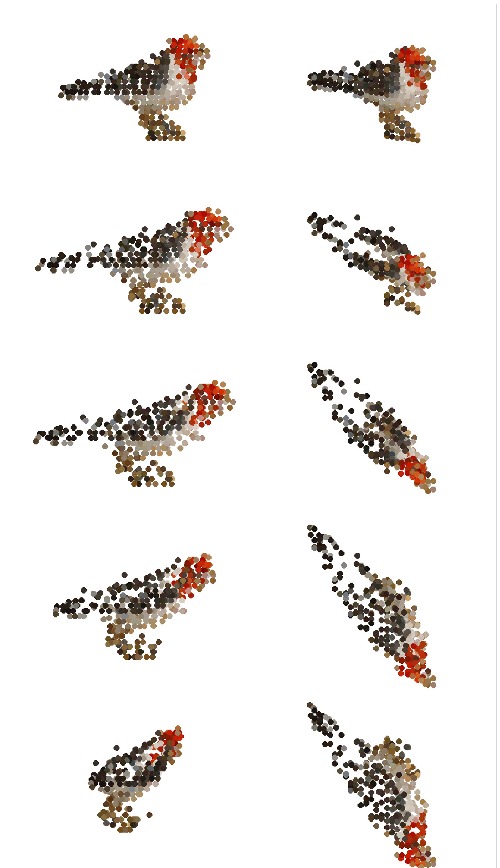} & 
\includegraphics[width = 0.1\textwidth, keepaspectratio = true]{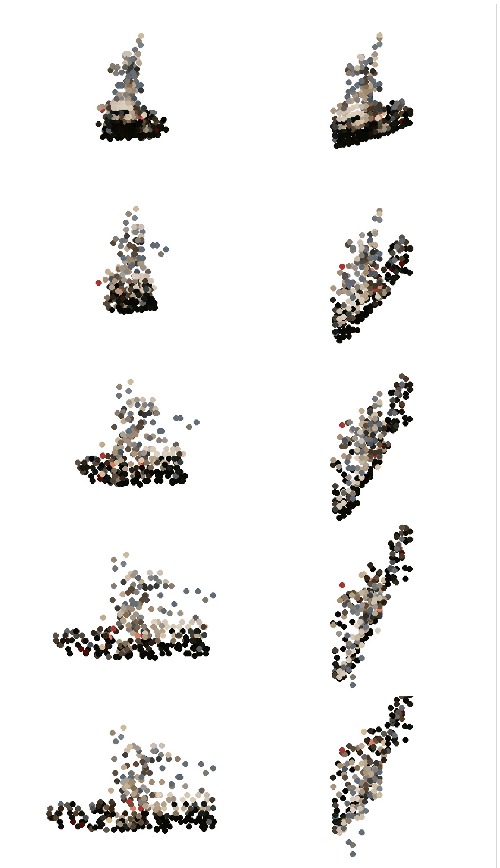} &
\includegraphics[width = 0.1\textwidth, keepaspectratio = true]{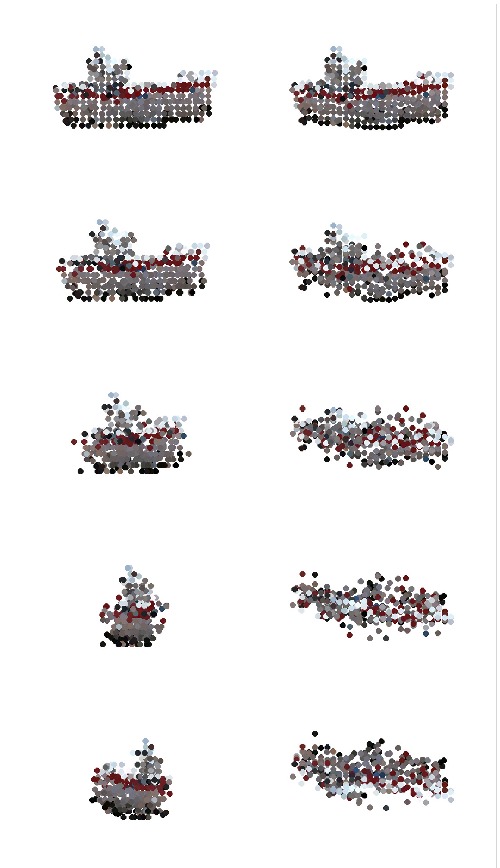} & 
\includegraphics[width = 0.1\textwidth, keepaspectratio = true]{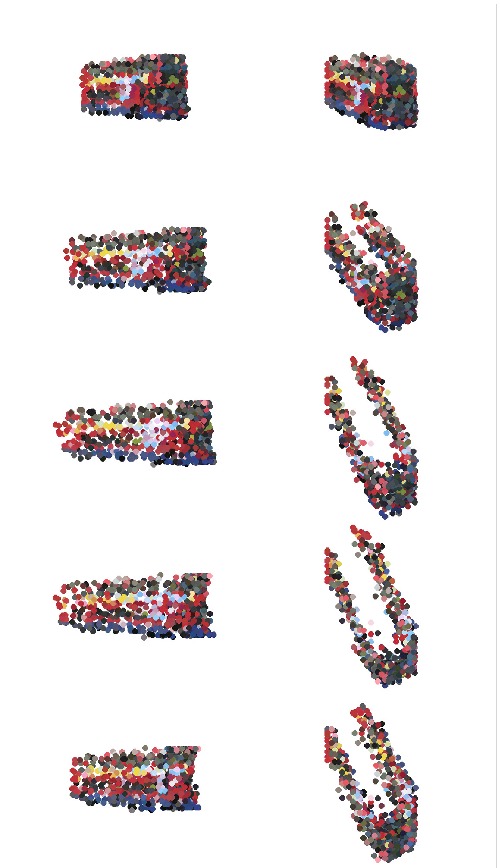} &
\includegraphics[width = 0.1\textwidth, keepaspectratio = true]{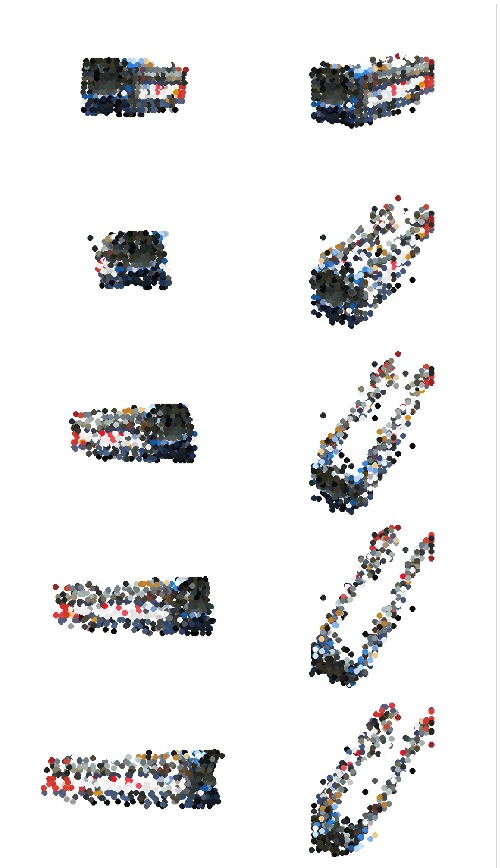} \\
\includegraphics[width = 0.1\textwidth, keepaspectratio = true]{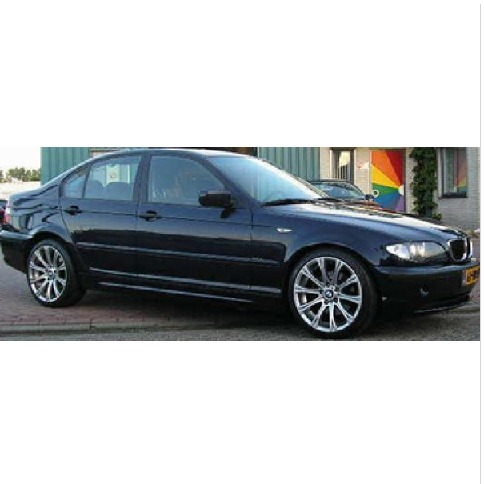} &
\includegraphics[width = 0.1\textwidth, keepaspectratio = true]{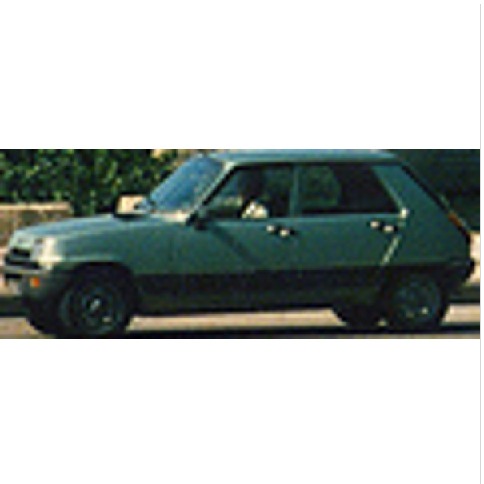} &
\includegraphics[width = 0.1\textwidth, keepaspectratio = true]{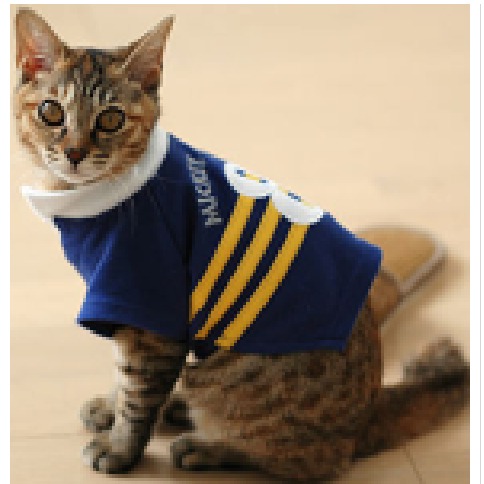} &
\includegraphics[width = 0.1\textwidth, keepaspectratio = true]{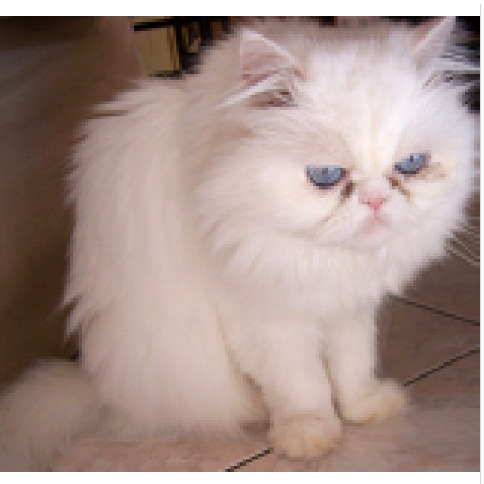} &
\includegraphics[width = 0.1\textwidth, keepaspectratio = true]{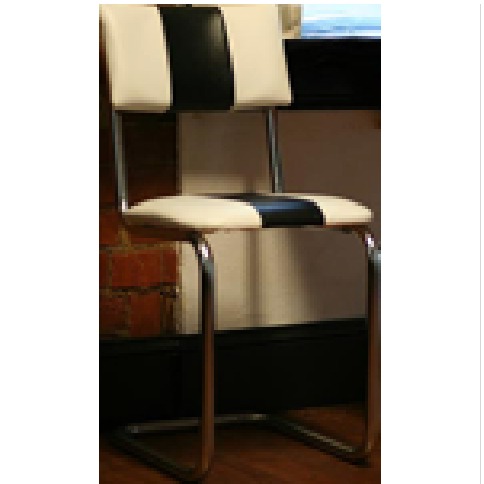} &
\includegraphics[width = 0.1\textwidth, keepaspectratio = true]{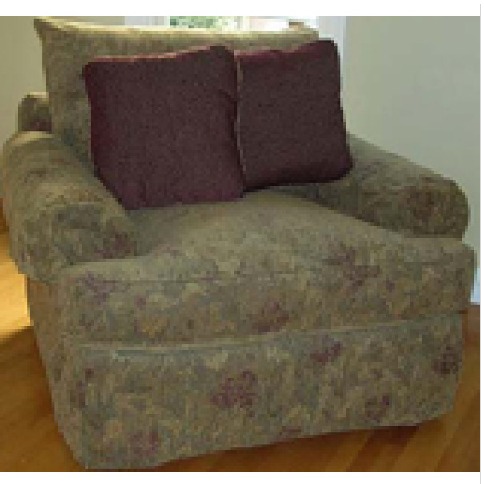} &
\includegraphics[width = 0.1\textwidth, keepaspectratio = true]{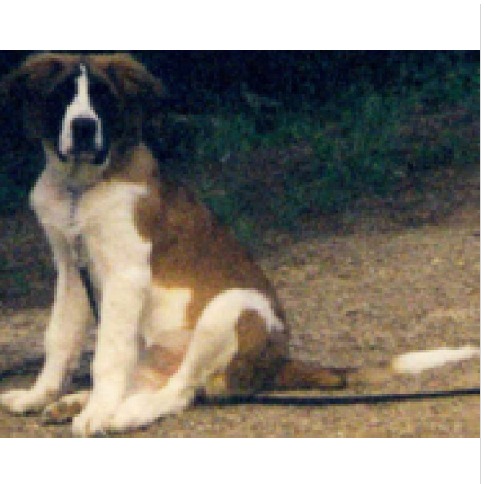} &
\includegraphics[width = 0.1\textwidth, keepaspectratio = true]{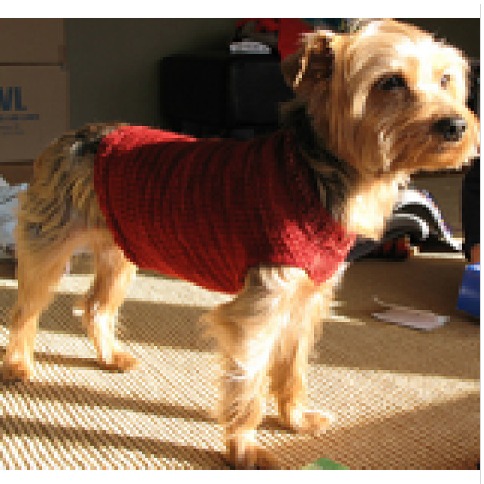} &
\includegraphics[width = 0.1\textwidth, keepaspectratio = true]{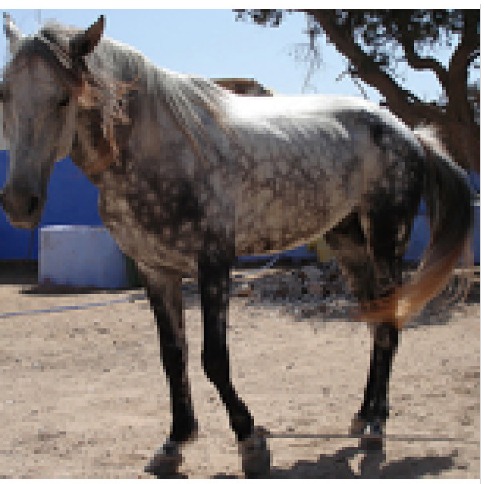} &
\includegraphics[width = 0.1\textwidth, keepaspectratio = true]{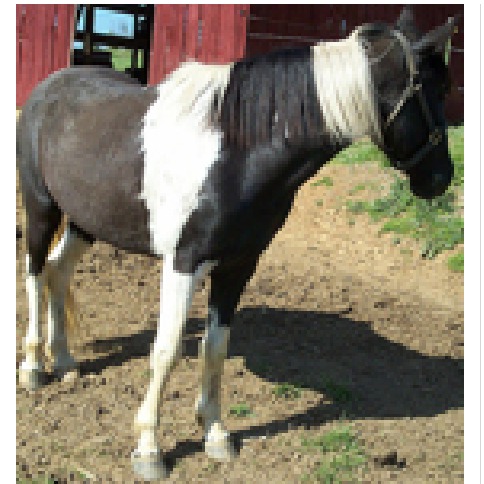} \\
\includegraphics[width = 0.1\textwidth, keepaspectratio = true]{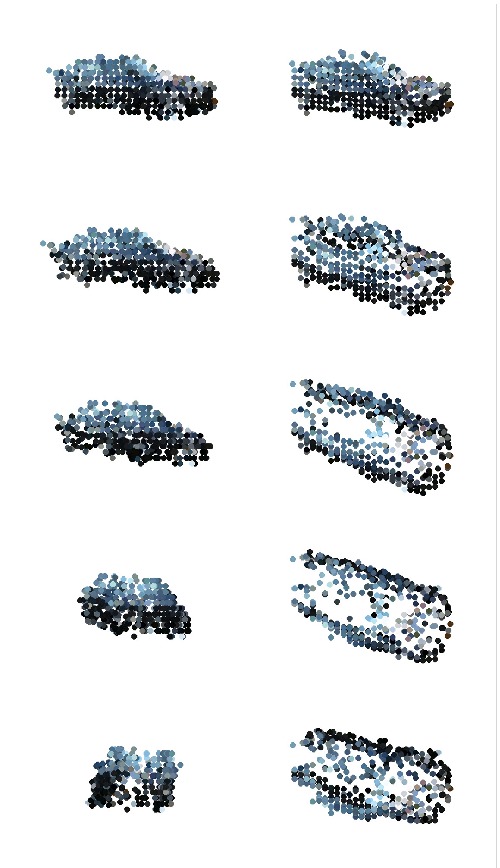} &
\includegraphics[width = 0.1\textwidth, keepaspectratio = true]{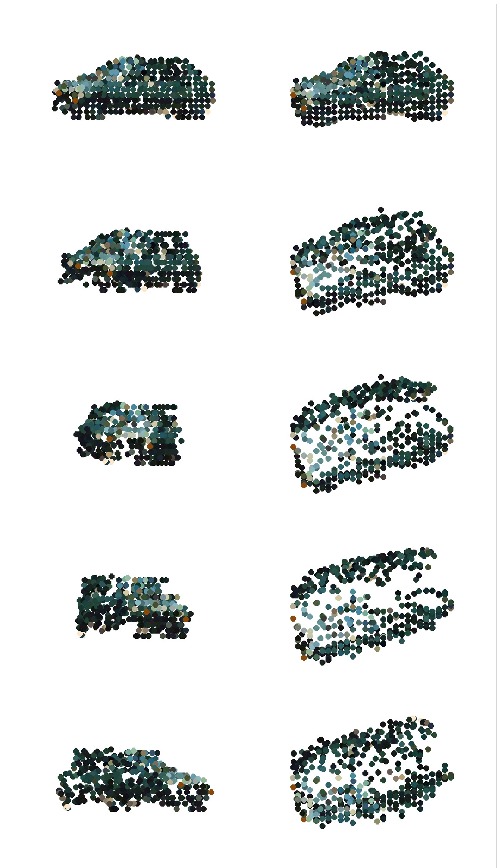} &
\includegraphics[width = 0.1\textwidth, keepaspectratio = true]{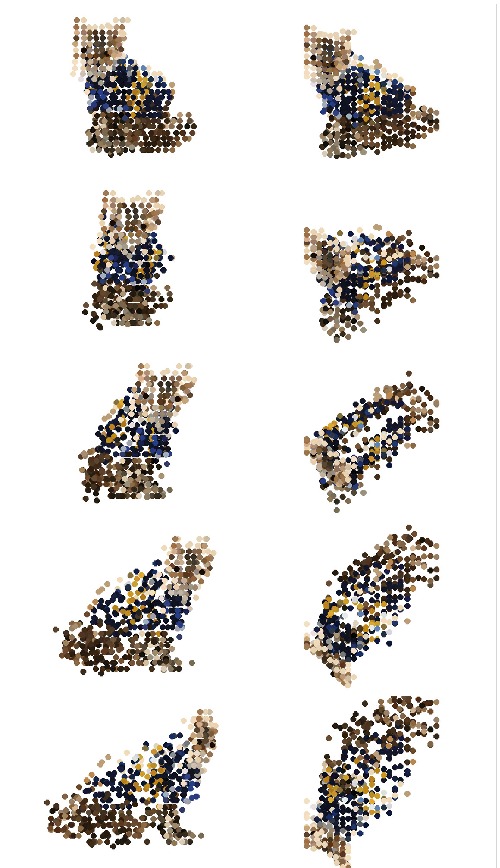} &
\includegraphics[width = 0.1\textwidth, keepaspectratio = true]{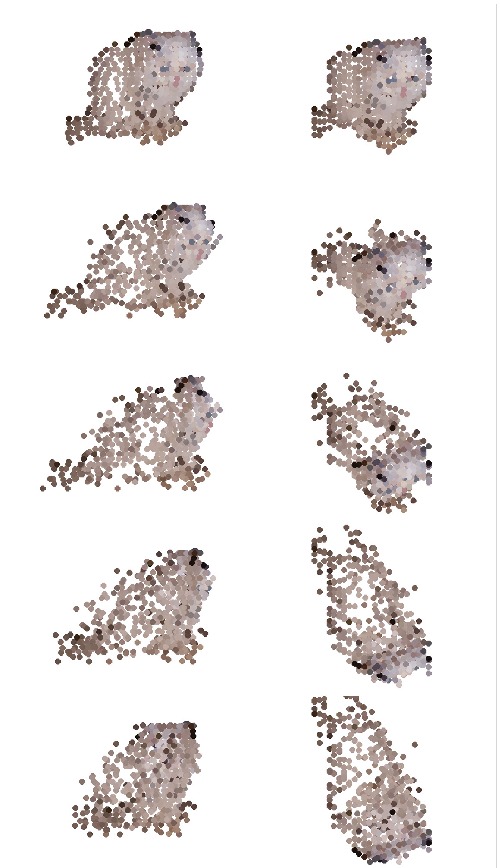} &
\includegraphics[width = 0.1\textwidth, keepaspectratio = true]{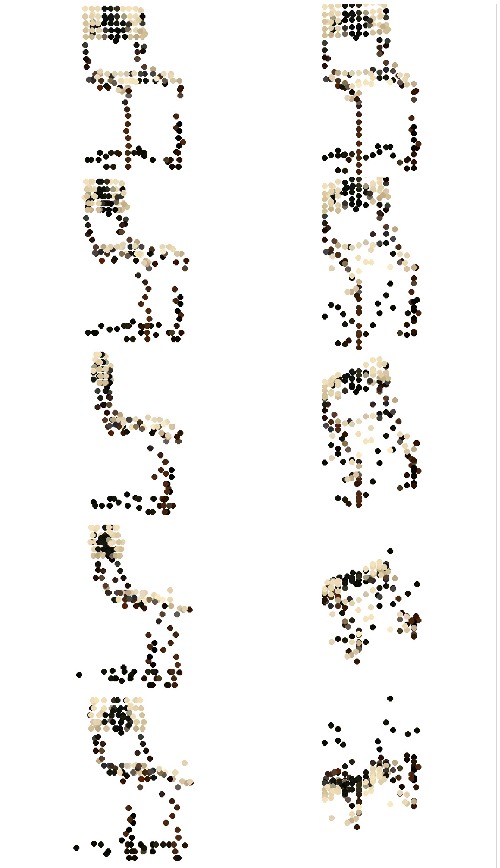} &
\includegraphics[width = 0.1\textwidth, keepaspectratio = true]{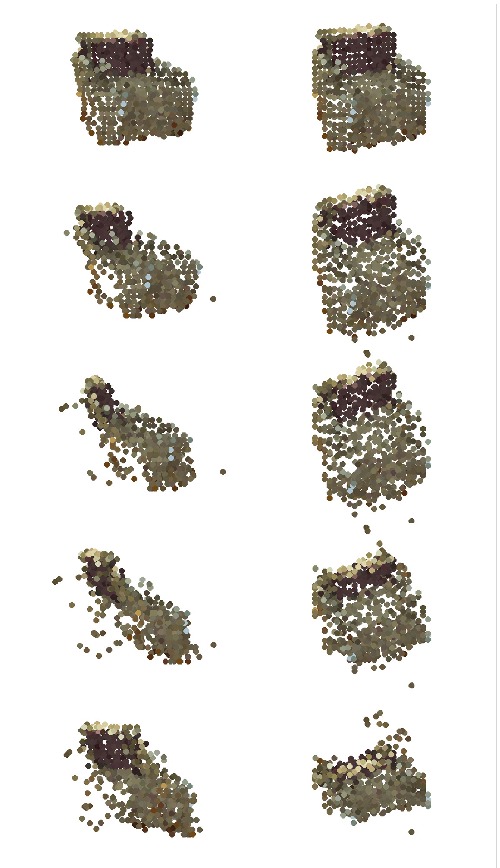} &
\includegraphics[width = 0.1\textwidth, keepaspectratio = true]{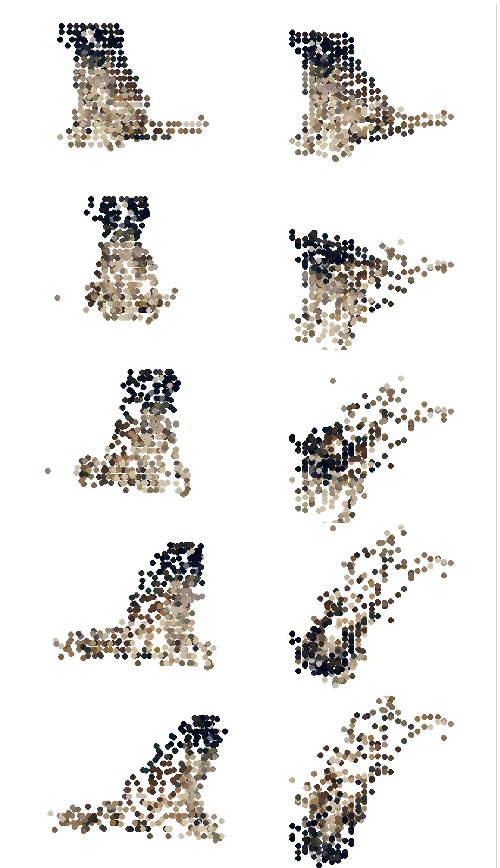} &
\includegraphics[width = 0.1\textwidth, keepaspectratio = true]{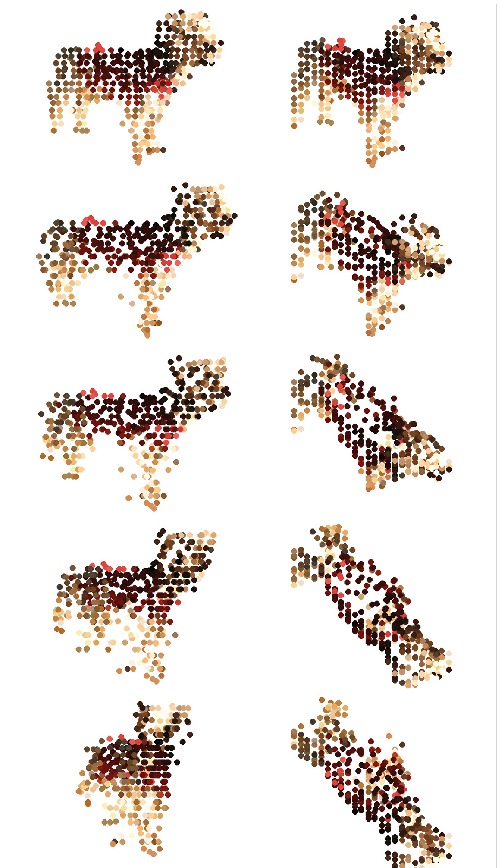} &
\includegraphics[width = 0.1\textwidth, keepaspectratio = true]{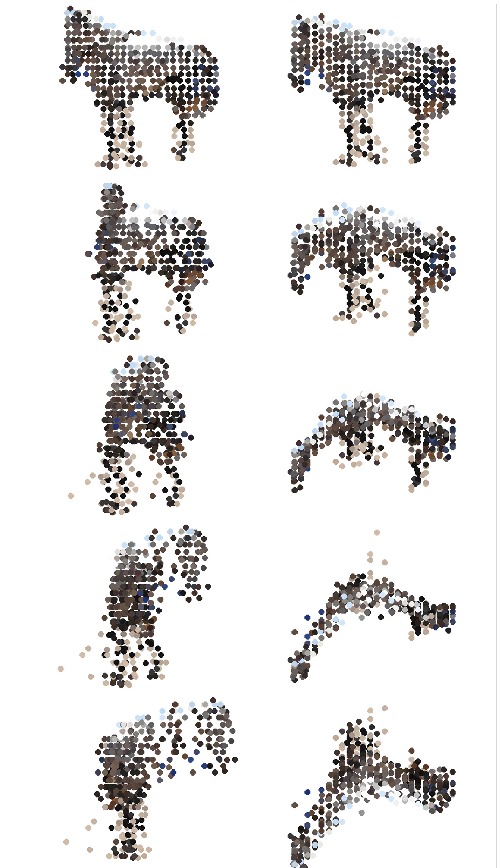} &
\includegraphics[width = 0.1\textwidth, keepaspectratio = true]{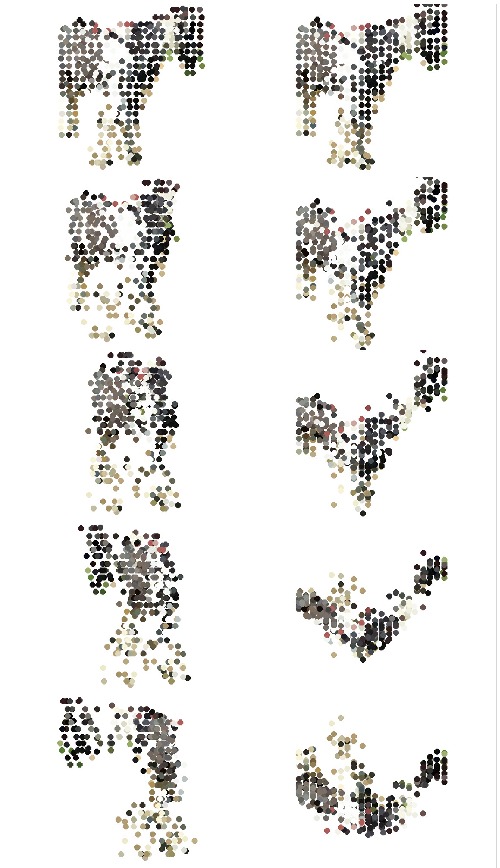} \\
\includegraphics[width = 0.1\textwidth, keepaspectratio = true]{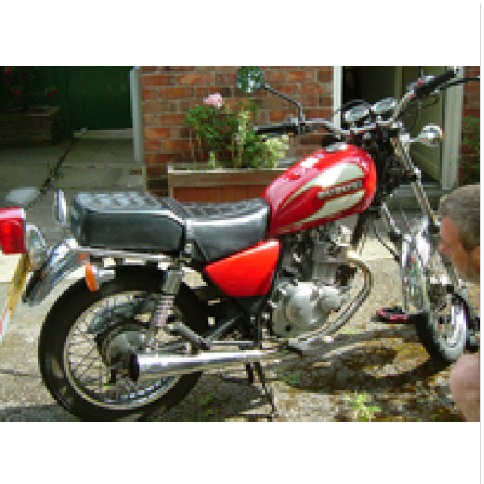} &
\includegraphics[width = 0.1\textwidth, keepaspectratio = true]{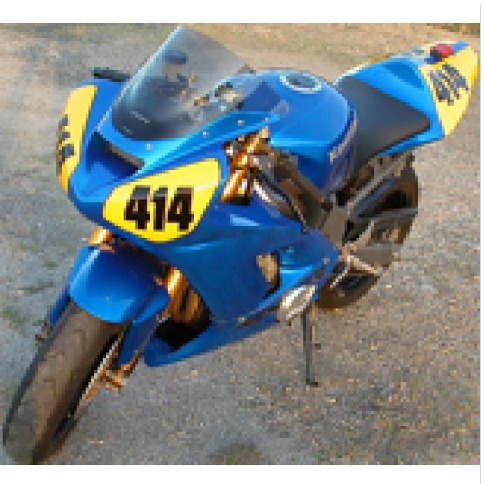} &
\includegraphics[width = 0.1\textwidth, keepaspectratio = true]{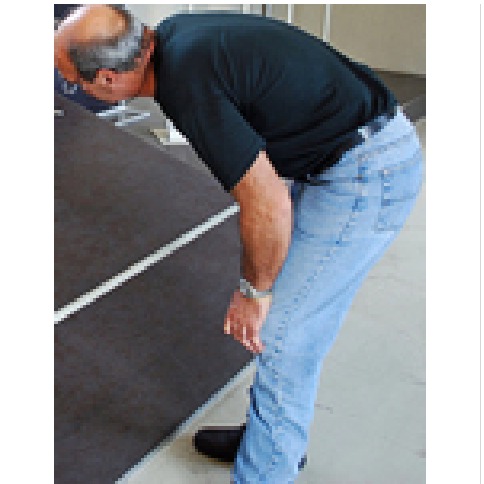} &
\includegraphics[width = 0.1\textwidth, keepaspectratio = true]{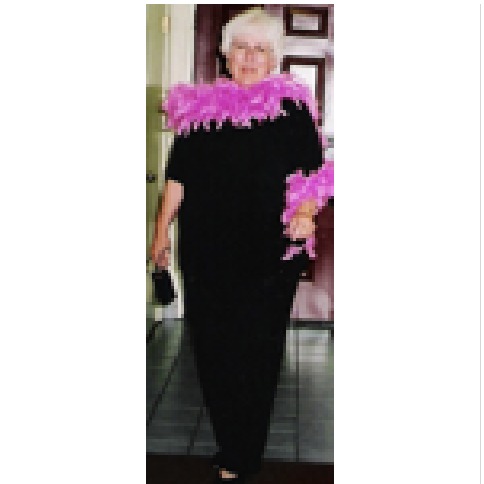} &
\includegraphics[width = 0.1\textwidth, keepaspectratio = true]{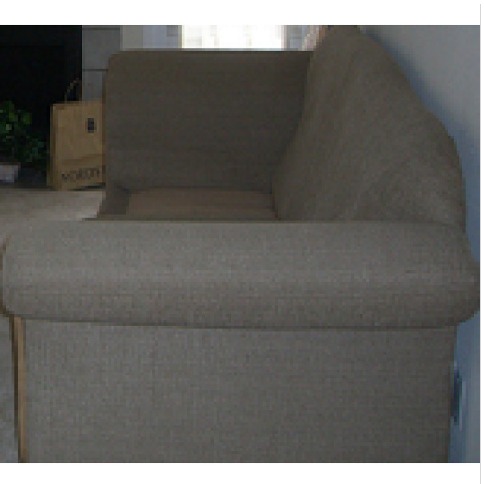} &
\includegraphics[width = 0.1\textwidth, keepaspectratio = true]{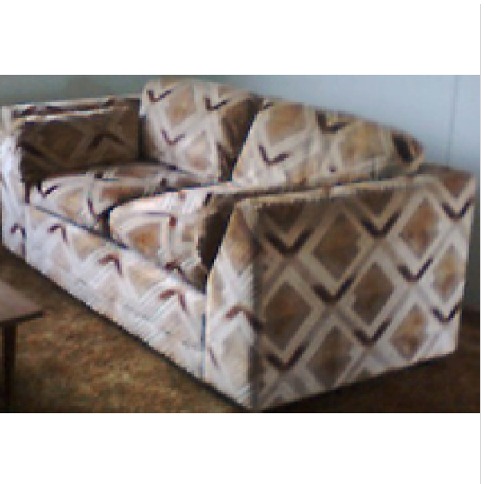} &
\includegraphics[width = 0.1\textwidth, keepaspectratio = true]{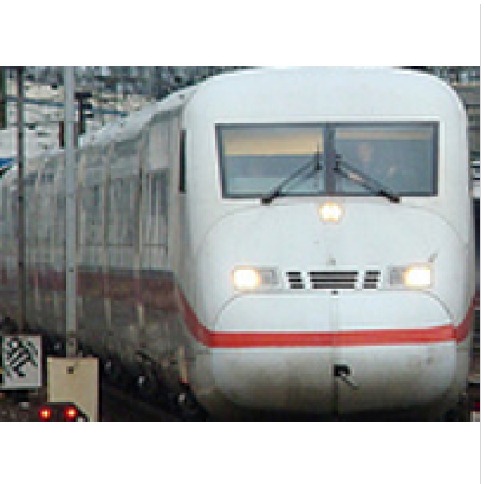} &
\includegraphics[width = 0.1\textwidth, keepaspectratio = true]{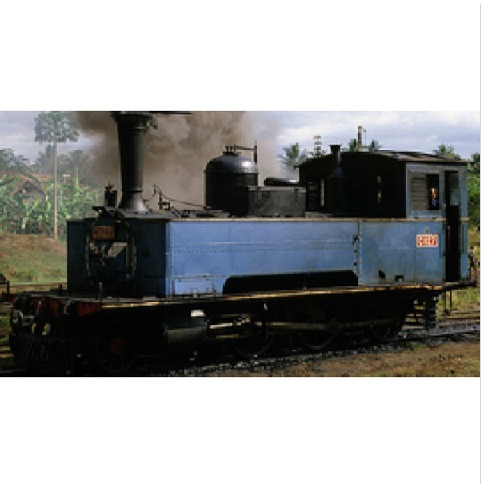} &
\includegraphics[width = 0.1\textwidth, keepaspectratio = true]{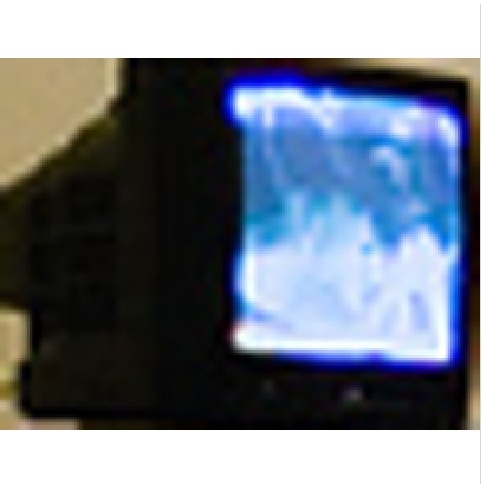} &
\includegraphics[width = 0.1\textwidth, keepaspectratio = true]{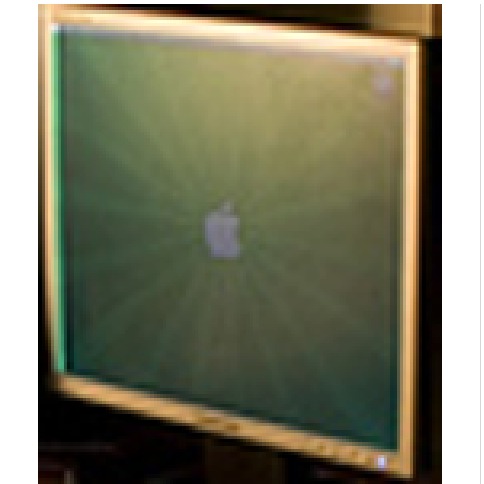} \\
\includegraphics[width = 0.1\textwidth, keepaspectratio = true]{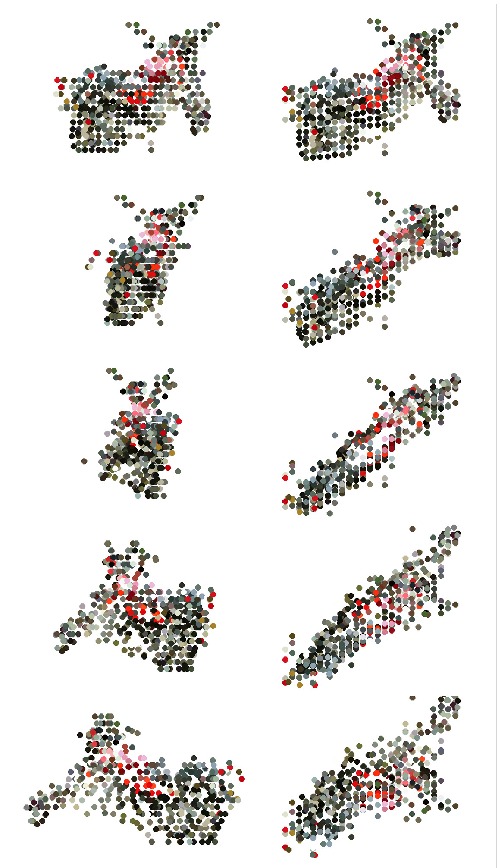} &
\includegraphics[width = 0.1\textwidth, keepaspectratio = true]{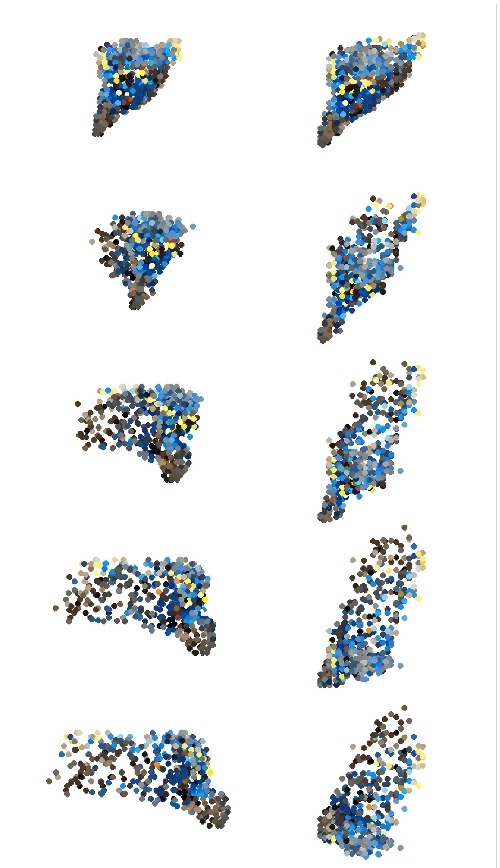} &
\includegraphics[width = 0.1\textwidth, keepaspectratio = true]{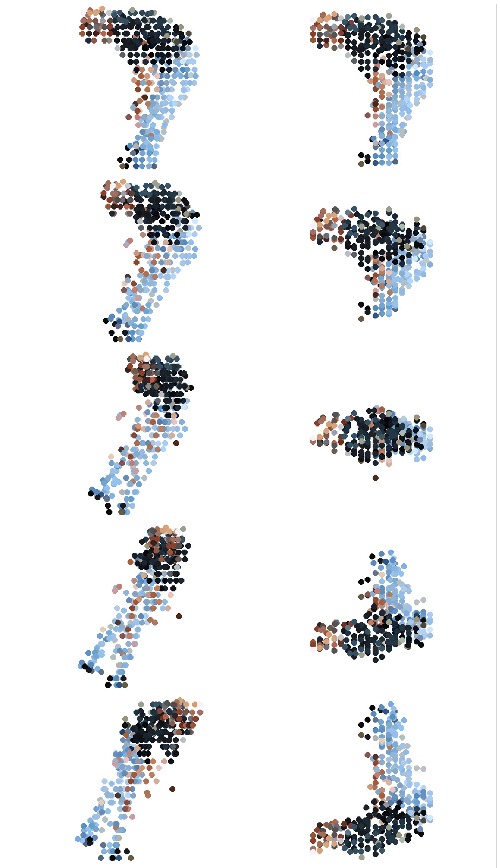} &
\includegraphics[width = 0.1\textwidth, keepaspectratio = true]{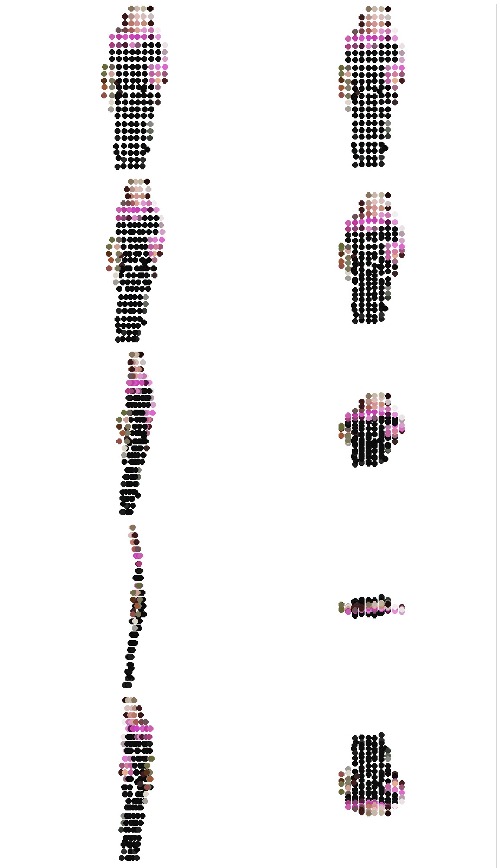} &
\includegraphics[width = 0.1\textwidth, keepaspectratio = true]{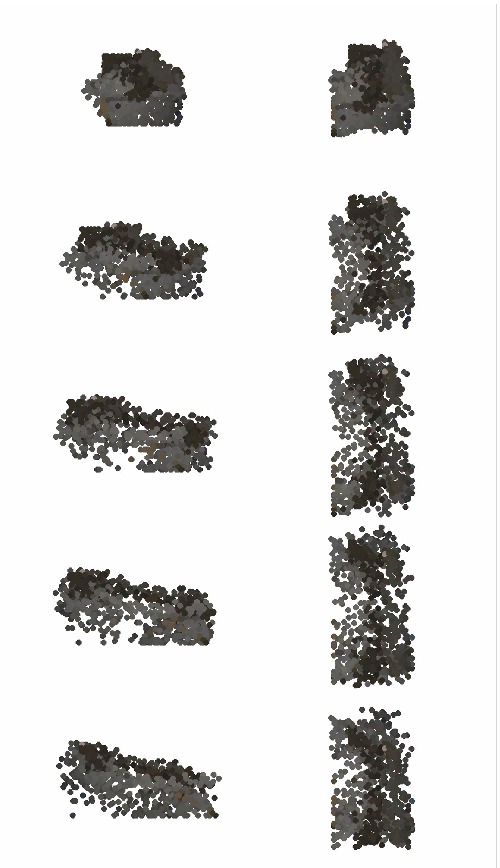} &
\includegraphics[width = 0.1\textwidth, keepaspectratio = true]{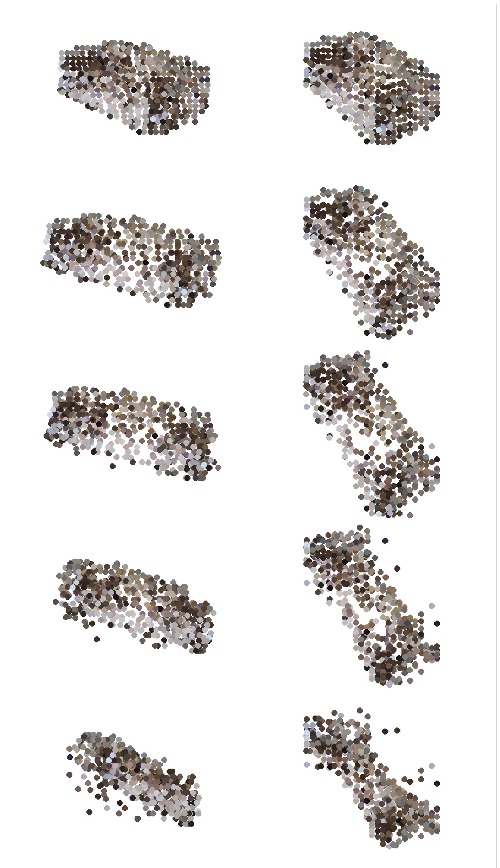} &
\includegraphics[width = 0.1\textwidth, keepaspectratio = true]{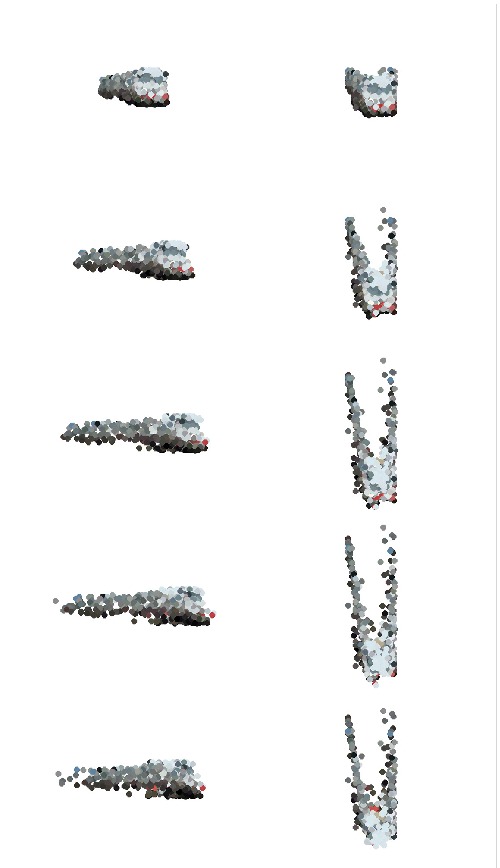} &
\includegraphics[width = 0.1\textwidth, keepaspectratio = true]{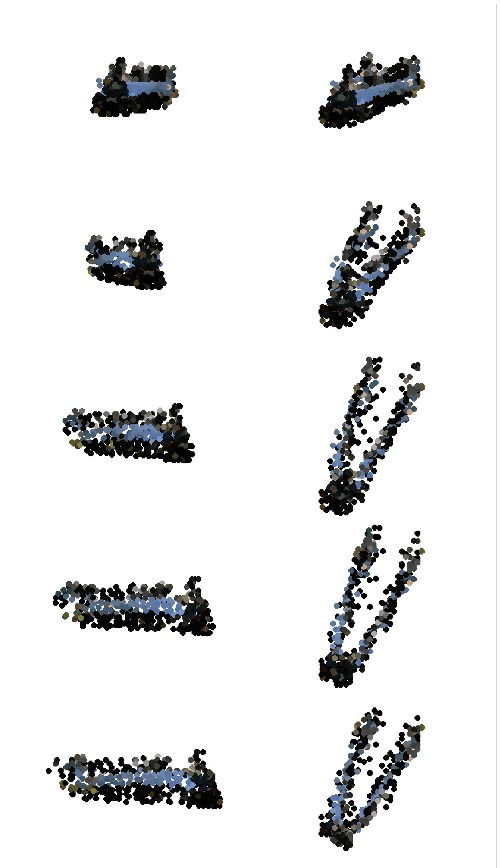} &
\includegraphics[width = 0.1\textwidth, keepaspectratio = true]{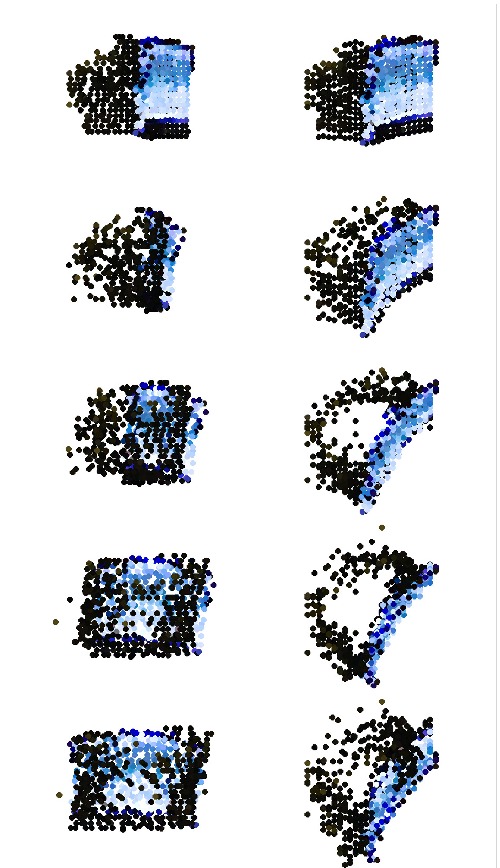} &
\includegraphics[width = 0.1\textwidth, keepaspectratio = true]{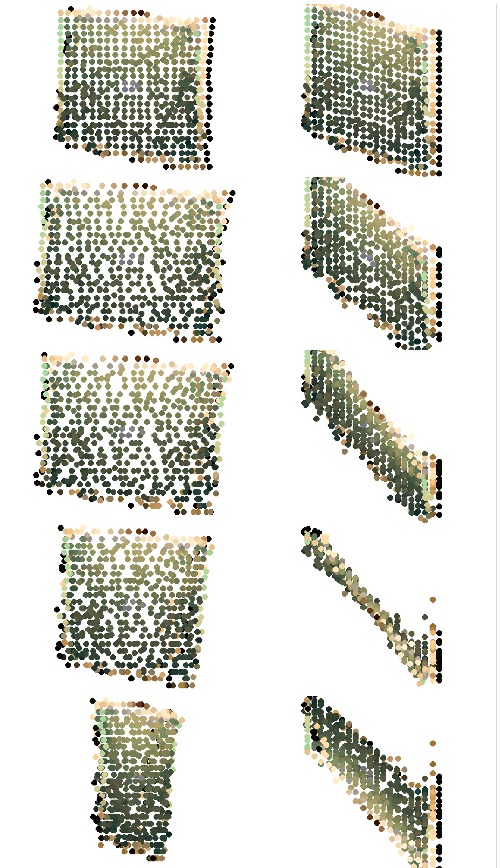} \\
\end{tabular}
\caption{\label{fig:reconstructions} Example reconstructions produced by VVN on 15 PASCAL VOC categories. The first column below each image shows shapes from increasing camera azimuths, the second from different elevations as in fig. \ref{fig1}. We do not show two classes due to lack of space: cows, which we shown in fig. \ref{fig1} and sheep, which are reconstructed with quality similar to cats and dogs. These reconstructions are available as a video online at \url{http://youtu.be/JfDJji5sYXE}. Reconstruction is quite successful for most classes, including most animal categories except horses, that are mostly flat and consistently get a long, tilted neck. Boat is not very good either, perhaps because due to extreme intra-class variation and aeroplane and birds wings are challenging to capture accurately when seen from the side. The person class is not aided much by bilateral symmetry, except for side views, and their reconstructions feel less 3D than for most other classes.} 
\end{figure*}

Reconstructions for all considered classes are shown in fig. \ref{fig:reconstructions}, assuming ground truth object segmentation and viewpoints from \cite{vicente2014reconstructing}. Inlier extrapolation helped visibly in many cases, especially for the tv/monitors class which completely failed without it, becoming curved shapes not unlike Dali clocks. Highly accurate shapes are obtained for most classes, the clearest exception being horses, seemingly due to noise in the cameras used. See the caption for additional comments. There is no existing dataset for evaluating this task, also because there are few, if any, methods developed for it, so we will simply make available all our reconstructions on the internet for anyone to evaluate. For now a selection is available on youtube at: \url{http://youtu.be/JfDJji5sYXE}. The method with closest capabilities that we aware of is \cite{vicente2014reconstructing}, but, while it produces a full mesh, it uses only silhouette information in a visual hull framework, e.g. image information is ignored and no correspondences are used, rendering it unable to deal with concavities, for instance.

\section{Conclusions}
\label{sec:conclusion} 

We have introduced a framework for shape reconstruction from a single image of a target object, using structure from motion on virtual views computed from a collection of training images of other objects from the same class. We proposed a novel method for 2D alignment that builds a network over the image collection in order to achieve robustness across wide viewpoint variation. We have also developed techniques to increase the robustness and specificity of factorization-based rigid structure from motion using virtual views and obtained stable and accurate reconstructions of challenging objects with diverse shapes. The ability to reconstruct from one image opens many new avenues for both structure from motion and recognition.

\vspace{3mm}
\noindent \textbf{Acknowledgements.} We would like to thank Sara Vicente for helpful comments. This work was supported in part by NSF Award IIS-1212798 and ONR MURI - N00014-10-1-0933. Jo\~{a}o Carreira was supported by the Portuguese Science Foundation, FCT, under grant SFRH/BPD/84194/2012 and Shubham Tulsiani was supported by the Berkeley fellowship. We gratefully acknowledge NVIDIA corporation for the donation of Tesla GPUs for this research.
{\small
\bibliographystyle{ieee}
\bibliography{all_refs}

\begin{thebibliography}{10}\itemsep=-1pt

\bibitem{aguiar2003rank}
P.~M. Aguiar and J.~M. Moura.
\newblock Rank 1 weighted factorization for 3d structure recovery: algorithms
  and performance analysis.
\newblock {\em Pattern Analysis and Machine Intelligence, IEEE Transactions
  on}, 25(9):1134--1149, 2003.

\bibitem{bao2013dense}
S.~Y. Bao, M.~Chandraker, Y.~Lin, and S.~Savarese.
\newblock Dense object reconstruction with semantic priors.
\newblock In {\em Computer Vision and Pattern Recognition (CVPR), 2013 IEEE
  Conference on}, pages 1264--1271. IEEE, 2013.

\bibitem{bao2011semantic}
S.~Y. Bao and S.~Savarese.
\newblock Semantic structure from motion.
\newblock In {\em Computer Vision and Pattern Recognition (CVPR), 2011 IEEE
  Conference on}, pages 2025--2032. IEEE, 2011.

\bibitem{barron2012shape}
J.~T. Barron and J.~Malik.
\newblock Shape, albedo, and illumination from a single image of an unknown
  object.
\newblock In {\em Computer Vision and Pattern Recognition (CVPR), 2012 IEEE
  Conference on}, pages 334--341. IEEE, 2012.

\bibitem{belhumeur2011localizing}
P.~N. Belhumeur, D.~W. Jacobs, D.~Kriegman, and N.~Kumar.
\newblock Localizing parts of faces using a consensus of exemplars.
\newblock In {\em CVPR}, pages 545--552. IEEE, 2011.

\bibitem{belongie2002shape}
S.~Belongie, J.~Malik, and J.~Puzicha.
\newblock Shape matching and object recognition using shape contexts.
\newblock {\em Pattern Analysis and Machine Intelligence, IEEE Transactions
  on}, 24(4):509--522, 2002.

\bibitem{boddeti2013correlation}
V.~N. Boddeti, T.~Kanade, and B.~Kumar.
\newblock Correlation filters for object alignment.
\newblock In {\em Computer Vision and Pattern Recognition (CVPR), 2013 IEEE
  Conference on}, pages 2291--2298. IEEE, 2013.

\bibitem{bookstein1997morphometric}
F.~L. Bookstein.
\newblock {\em Morphometric tools for landmark data: geometry and biology}.
\newblock Cambridge University Press, 1997.

\bibitem{bregler2000recovering}
C.~Bregler, A.~Hertzmann, and H.~Biermann.
\newblock Recovering non-rigid 3d shape from image streams.
\newblock In {\em Computer Vision and Pattern Recognition, 2000. Proceedings.
  IEEE Conference on}, volume~2, pages 690--696. IEEE, 2000.

\bibitem{carreira2012semantic}
J.~Carreira, R.~Caseiro, J.~Batista, and C.~Sminchisescu.
\newblock Semantic segmentation with second-order pooling.
\newblock In {\em Computer Vision--ECCV 2012}, pages 430--443. Springer, 2012.

\bibitem{cashman2013shape}
T.~J. Cashman and A.~W. Fitzgibbon.
\newblock What shape are dolphins? building 3d morphable models from 2d images.
\newblock {\em Pattern Analysis and Machine Intelligence, IEEE Transactions
  on}, 35(1):232--244, 2013.

\bibitem{cootes1995active}
T.~F. Cootes, C.~J. Taylor, D.~H. Cooper, and J.~Graham.
\newblock Active shape models-their training and application.
\newblock {\em CVIU}, 61(1):38--59, 1995.

\bibitem{crandall2011discrete}
D.~Crandall, A.~Owens, N.~Snavely, and D.~Huttenlocher.
\newblock Discrete-continuous optimization for large-scale structure from
  motion.
\newblock In {\em Computer Vision and Pattern Recognition (CVPR), 2011 IEEE
  Conference on}, pages 3001--3008. IEEE, 2011.

\bibitem{dame2013dense}
A.~Dame, V.~A. Prisacariu, C.~Y. Ren, and I.~Reid.
\newblock Dense reconstruction using 3d object shape priors.
\newblock In {\em Computer Vision and Pattern Recognition (CVPR), 2013 IEEE
  Conference on}, pages 1288--1295. IEEE, 2013.

\bibitem{de2003framework}
F.~De~La~Torre and M.~J. Black.
\newblock A framework for robust subspace learning.
\newblock {\em International Journal of Computer Vision}, 54(1-3):117--142,
  2003.

\bibitem{eigen2014depth}
D.~Eigen, C.~Puhrsch, and R.~Fergus.
\newblock Depth map prediction from a single image using a multi-scale deep
  network.
\newblock {\em arXiv preprint arXiv:1406.2283}, 2014.

\bibitem{everingham2010pascal}
M.~Everingham, L.~Van~Gool, C.~K. Williams, J.~Winn, and A.~Zisserman.
\newblock The pascal visual object classes (voc) challenge.
\newblock {\em International journal of computer vision}, 88(2):303--338, 2010.

\bibitem{fredman1987fibonacci}
M.~L. Fredman and R.~E. Tarjan.
\newblock Fibonacci heaps and their uses in improved network optimization
  algorithms.
\newblock {\em Journal of the ACM (JACM)}, 34(3):596--615, 1987.

\bibitem{furukawa2010accurate}
Y.~Furukawa and J.~Ponce.
\newblock Accurate, dense, and robust multiview stereopsis.
\newblock {\em Pattern Analysis and Machine Intelligence, IEEE Transactions
  on}, 32(8):1362--1376, 2010.

\bibitem{hane2014class}
C.~Hane, N.~Savinov, and M.~Pollefeys.
\newblock Class specific 3d object shape priors using surface normals.
\newblock In {\em Computer Vision and Pattern Recognition (CVPR), 2014 IEEE
  Conference on}, pages 652--659. IEEE, 2014.

\bibitem{hariharan2011semantic}
B.~Hariharan, P.~Arbel{\'a}ez, L.~Bourdev, S.~Maji, and J.~Malik.
\newblock Semantic contours from inverse detectors.
\newblock In {\em Computer Vision (ICCV), 2011 IEEE International Conference
  on}, pages 991--998. IEEE, 2011.

\bibitem{hariharan2014simultaneous}
B.~Hariharan, P.~Arbel{\'a}ez, R.~Girshick, and J.~Malik.
\newblock Simultaneous detection and segmentation.
\newblock In {\em Computer Vision--ECCV 2014}, pages 297--312. Springer, 2014.

\bibitem{hartley2003multiple}
R.~Hartley and A.~Zisserman.
\newblock {\em Multiple view geometry in computer vision}.
\newblock Cambridge university press, 2003.

\bibitem{hassner2013single}
T.~Hassner and R.~Basri.
\newblock Single view depth estimation from examples.
\newblock {\em arXiv preprint arXiv:1304.3915}, 2013.

\bibitem{hejrati2014analysis}
M.~Hejrati and D.~Ramanan.
\newblock Analysis by synthesis: 3d object recognition by object
  reconstruction.
\newblock In {\em Computer Vision and Pattern Recognition (CVPR), 2014 IEEE
  Conference on}, pages 2449--2456. IEEE, 2014.

\bibitem{hoiem2007recovering}
D.~Hoiem, A.~A. Efros, and M.~Hebert.
\newblock Recovering surface layout from an image.
\newblock {\em International Journal of Computer Vision}, 75(1):151--172, 2007.

\bibitem{irani2000factorization}
M.~Irani and P.~Anandan.
\newblock Factorization with uncertainty.
\newblock In {\em Computer Vision-ECCV 2000}, pages 539--553. Springer, 2000.

\bibitem{jansson1973visual}
G.~Jansson and G.~Johansson.
\newblock Visual perception of bending motion.
\newblock {\em Perception}, 2(3):321--326, 1973.

\bibitem{karsch2013boundary}
K.~Karsch, Z.~Liao, J.~Rock, J.~T. Barron, and D.~Hoiem.
\newblock Boundary cues for 3d object shape recovery.
\newblock In {\em Computer Vision and Pattern Recognition (CVPR), 2013 IEEE
  Conference on}, pages 2163--2170. IEEE, 2013.

\bibitem{Karsch:TPAMI:14}
K.~Karsch, C.~Liu, and S.~B. Kang.
\newblock Depthtransfer: Depth extraction from video using non-parametric
  sampling.
\newblock {\em Pattern Analysis and Machine Intelligence, IEEE Transactions
  on}, 2014.

\bibitem{kemelmacher2013internet}
I.~Kemelmacher-Shlizerman.
\newblock Internet based morphable model.
\newblock In {\em Computer Vision (ICCV), 2013 IEEE International Conference
  on}, pages 3256--3263. IEEE, 2013.

\bibitem{kemelmacher2011face}
I.~Kemelmacher-Shlizerman and S.~M. Seitz.
\newblock Face reconstruction in the wild.
\newblock In {\em Computer Vision (ICCV), 2011 IEEE International Conference
  on}, pages 1746--1753. IEEE, 2011.

\bibitem{kim2013deformable}
J.~Kim, C.~Liu, F.~Sha, and K.~Grauman.
\newblock Deformable spatial pyramid matching for fast dense correspondences.
\newblock In {\em Computer Vision and Pattern Recognition (CVPR), 2013 IEEE
  Conference on}, pages 2307--2314. IEEE, 2013.

\bibitem{KrauseStarkDengFei-Fei_3DRR2013}
J.~Krause, M.~Stark, J.~Deng, and L.~Fei-Fei.
\newblock 3d object representations for fine-grained categorization.
\newblock In {\em 4th International IEEE Workshop on 3D Representation and
  Recognition (3dRR-13)}, Sydney, Australia, 2013.

\bibitem{krizhevsky2012imagenet}
A.~Krizhevsky, I.~Sutskever, and G.~E. Hinton.
\newblock Imagenet classification with deep convolutional neural networks.
\newblock In {\em Advances in neural information processing systems}, pages
  1097--1105, 2012.

\bibitem{ladicky2014pulling}
L.~Ladicky, J.~Shi, and M.~Pollefeys.
\newblock Pulling things out of perspective.
\newblock In {\em Computer Vision and Pattern Recognition (CVPR), 2014 IEEE
  Conference on}, pages 89--96. IEEE, 2014.

\bibitem{liu2011sift}
C.~Liu, J.~Yuen, and A.~Torralba.
\newblock Sift flow: Dense correspondence across scenes and its applications.
\newblock {\em TPAMI}, 33(5):978--994, 2011.

\bibitem{malisiewicz2009beyond}
T.~Malisiewicz and A.~Efros.
\newblock Beyond categories: The visual memex model for reasoning about object
  relationships.
\newblock In {\em Advances in neural information processing systems}, pages
  1222--1230, 2009.

\bibitem{marques2009estimating}
M.~Marques and J.~Costeira.
\newblock Estimating 3d shape from degenerate sequences with missing data.
\newblock {\em Computer Vision and Image Understanding}, 113(2):261--272, 2009.

\bibitem{martins2012discriminative}
P.~Martins, R.~Caseiro, J.~F. Henriques, and J.~Batista.
\newblock Discriminative bayesian active shape models.
\newblock In {\em European Conference on Computer Vision}, 2012.

\bibitem{paladini2009factorization}
M.~Paladini, A.~Del~Bue, M.~Stosic, M.~Dodig, J.~Xavier, and L.~Agapito.
\newblock Factorization for non-rigid and articulated structure using metric
  projections.
\newblock In {\em Computer Vision and Pattern Recognition, 2009. CVPR 2009.
  IEEE Conference on}, pages 2898--2905. IEEE, 2009.

\bibitem{pepik2012teaching}
B.~Pepik, M.~Stark, P.~Gehler, and B.~Schiele.
\newblock Teaching 3d geometry to deformable part models.
\newblock In {\em Computer Vision and Pattern Recognition (CVPR), 2012 IEEE
  Conference on}, pages 3362--3369. IEEE, 2012.

\bibitem{rother2006cosegmentation}
C.~Rother, T.~Minka, A.~Blake, and V.~Kolmogorov.
\newblock Cosegmentation of image pairs by histogram matching-incorporating a
  global constraint into mrfs.
\newblock In {\em Computer Vision and Pattern Recognition, 2006 IEEE Computer
  Society Conference on}, volume~1, pages 993--1000. IEEE, 2006.

\bibitem{rubinstein2013unsupervised}
M.~Rubinstein, A.~Joulin, J.~Kopf, and C.~Liu.
\newblock Unsupervised joint object discovery and segmentation in internet
  images.
\newblock In {\em CVPR}, pages 1939--1946, 2013.

\bibitem{saxena2009make3d}
A.~Saxena, M.~Sun, and A.~Y. Ng.
\newblock Make3d: Learning 3d scene structure from a single still image.
\newblock {\em Pattern Analysis and Machine Intelligence, IEEE Transactions
  on}, 31(5):824--840, 2009.

\bibitem{sermanet2013overfeat}
P.~Sermanet, D.~Eigen, X.~Zhang, M.~Mathieu, R.~Fergus, and Y.~LeCun.
\newblock Overfeat: Integrated recognition, localization and detection using
  convolutional networks.
\newblock {\em arXiv preprint arXiv:1312.6229}, 2013.

\bibitem{snavely2006photo}
N.~Snavely, S.~M. Seitz, and R.~Szeliski.
\newblock Photo tourism: exploring photo collections in 3d.
\newblock {\em ACM transactions on graphics (TOG)}, 25(3):835--846, 2006.

\bibitem{srebro2003weighted}
N.~Srebro, T.~Jaakkola, et~al.
\newblock Weighted low-rank approximations.
\newblock In {\em ICML}, volume~3, pages 720--727, 2003.

\bibitem{su2014estimating}
H.~Su, Q.~Huang, N.~Mitra, Y.~Li, and L.~Guibas.
\newblock Estimating image depth using shape collection.
\newblock {\em Transaction of Graphics}, (Special Issue of SIGGRAPH 2014),
  2014.

\bibitem{su09}
H.~Su, M.~Sun, L.~Fei-Fei, and S.~Savarese.
\newblock Learning a dense multi-view representation for detection, viewpoint
  classification and synthesis of object categories.
\newblock In {\em CVPR}, pages 213--220, 2009.

\bibitem{tomasi1992shape}
C.~Tomasi and T.~Kanade.
\newblock Shape and motion from image streams under orthography: a
  factorization method.
\newblock {\em International Journal of Computer Vision}, 9(2):137--154, 1992.

\bibitem{ullman1983maximizing}
S.~Ullman.
\newblock Maximizing rigidity: the incremental recovery of 3-d structure from
  rigid and rubbery motion.
\newblock 1983.

\bibitem{vicente2014reconstructing}
S.~Vicente, J.~Carreira, L.~Agapito, and J.~Batista.
\newblock Reconstructing pascal voc.
\newblock In {\em Computer Vision and Pattern Recognition (CVPR), 2014 IEEE
  Conference on}, pages 41--48. IEEE, 2014.

\bibitem{xiang2014beyond}
Y.~Xiang, R.~Mottaghi, and S.~Savarese.
\newblock Beyond pascal: A benchmark for 3d object detection in the wild.
\newblock {\em IEEE Winter Conference on Applications of Computer Vision
  (WACV)}, 2014.

\bibitem{xu1995robust}
L.~Xu and A.~L. Yuille.
\newblock Robust principal component analysis by self-organizing rules based on
  statistical physics approach.
\newblock {\em Neural Networks, IEEE Transactions on}, 6(1):131--143, 1995.

\bibitem{yang2013articulated}
Y.~Yang and D.~Ramanan.
\newblock Articulated human detection with flexible mixtures-of-parts.
\newblock {\em Pattern Analysis and Machine Intelligence, IEEE Transactions
  on}, 35(12):2878--2890, 2013.

\end{thebibliography}
}




\pagebreak
\section*{Supplementary Material}

Additional automatic alignments (assuming only ground truth segmentations) for those classes not shown in fig. 8 of the main body of the paper can be found here. 

\begin{figure*}
\centering
\renewcommand{\arraystretch}{1}
\begin{tabular}{@{}c@{} c@{}}
\includegraphics[width=0.35\linewidth]{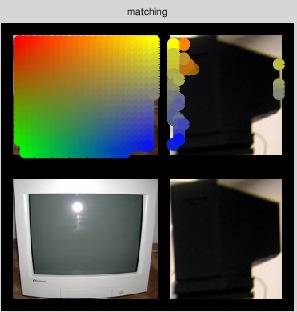} &
\includegraphics[width=0.35\linewidth]{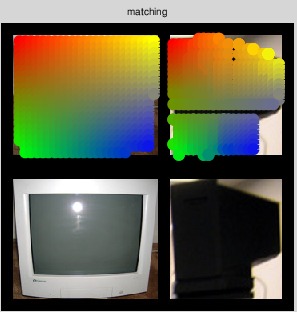} \\
\includegraphics[width=0.35\linewidth]{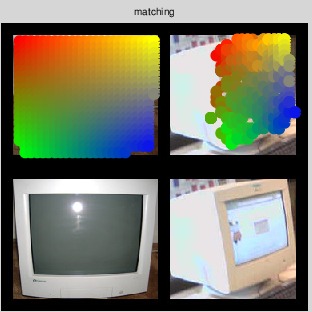} &
\includegraphics[width=0.35\linewidth]{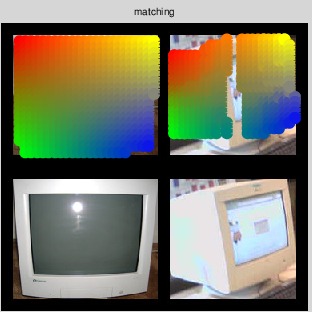} \\ 
\includegraphics[width=0.35\linewidth]{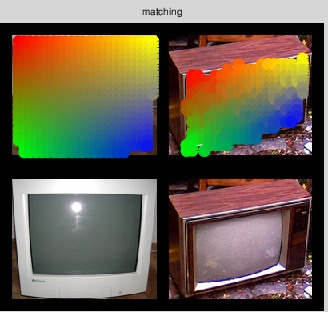} & 
\includegraphics[width=0.35\linewidth]{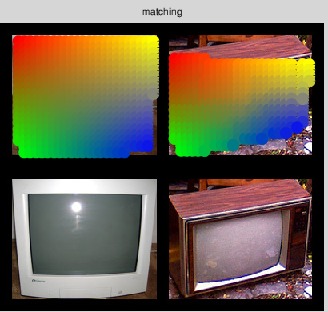} \\
\end{tabular}
\caption{\label{fig:alignments} Example tvmonitor alignments using our proposed network-based approach with automatic pose prediction (first two columns), named VVN,  and SIFTflow (last two columns), on the same grid of deep features and assuming correct figure-ground segmentation. Corresponding points are colored the same. VVN exploits class-specific knowledge and pose prediction to obtain resillience to viewpoint variation.}
\end{figure*}


\begin{figure*}
\centering
\renewcommand{\arraystretch}{1}
\begin{tabular}{@{}c@{} c@{}}
\includegraphics[width=0.35\linewidth]{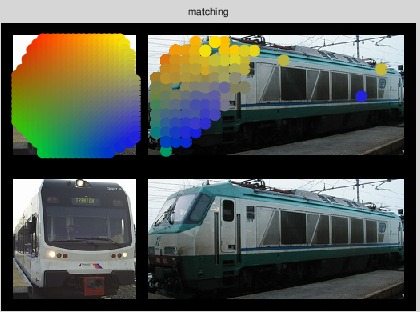} &
\includegraphics[width=0.35\linewidth]{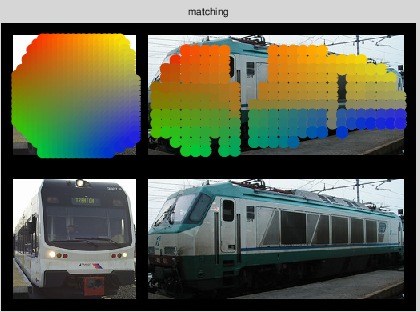} \\
\includegraphics[width=0.35\linewidth]{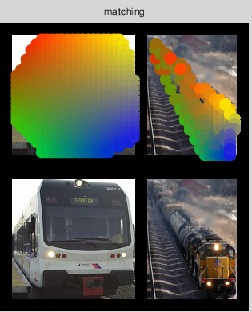} &
\includegraphics[width=0.35\linewidth]{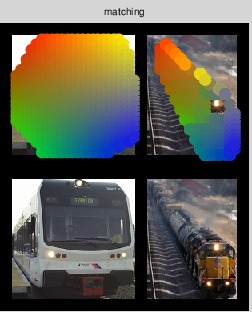} \\ 
\includegraphics[width=0.35\linewidth]{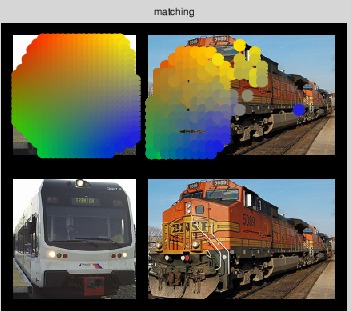} & 
\includegraphics[width=0.35\linewidth]{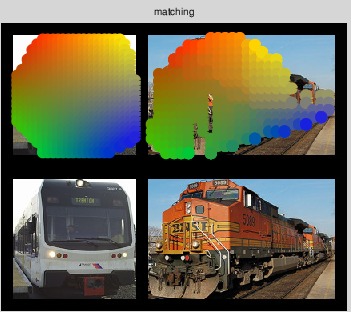} \\
\end{tabular}
\caption{\label{fig:alignments} Example train alignments. On the left the proposed method VVN, on the right SIFTflow.}
\end{figure*}


\begin{figure*}
\centering
\renewcommand{\arraystretch}{1}
\begin{tabular}{@{}c@{} c@{}}
\includegraphics[width=0.49\linewidth]{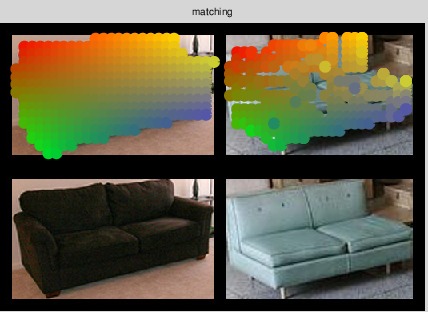} &
\includegraphics[width=0.49\linewidth]{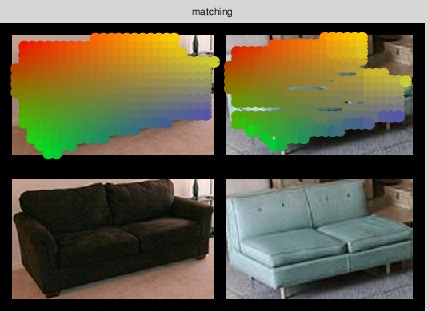} \\
\includegraphics[width=0.49\linewidth]{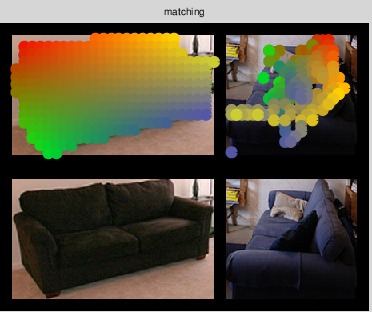} &
\includegraphics[width=0.49\linewidth]{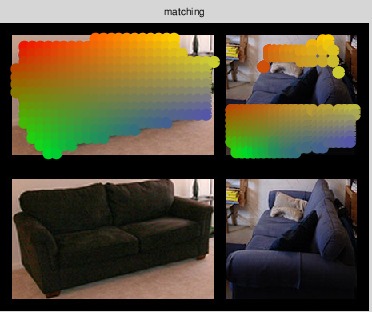} \\ 
\includegraphics[width=0.49\linewidth]{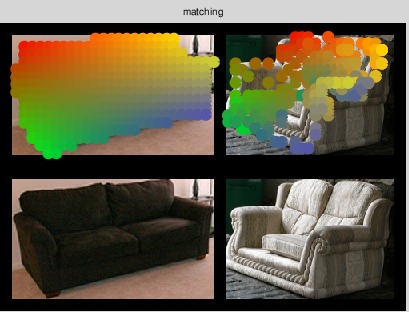} & 
\includegraphics[width=0.49\linewidth]{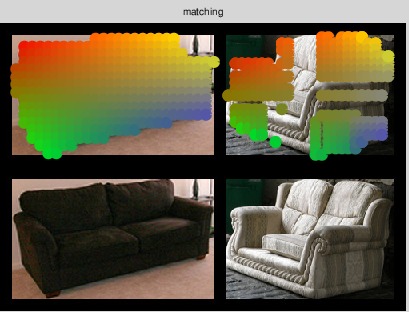} \\
\end{tabular}
\caption{\label{fig:alignments} Example sofa alignments. On the left the proposed method VVN, on the right SIFTflow.}
\end{figure*}


\begin{figure*}
\centering
\renewcommand{\arraystretch}{1}
\begin{tabular}{@{}c@{} c@{}}
\includegraphics[width=0.30\linewidth]{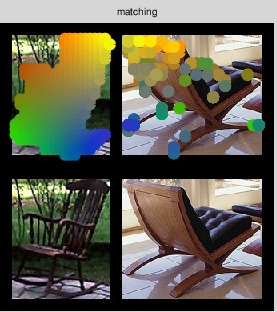} &
\includegraphics[width=0.30\linewidth]{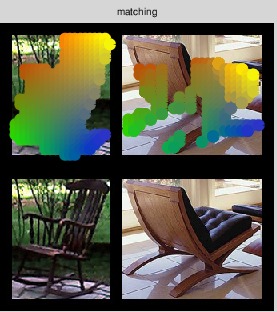} \\
\includegraphics[width=0.30\linewidth]{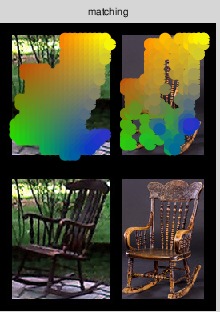} &
\includegraphics[width=0.30\linewidth]{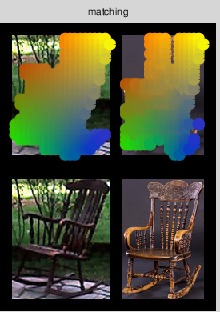} \\ 
\includegraphics[width=0.30\linewidth]{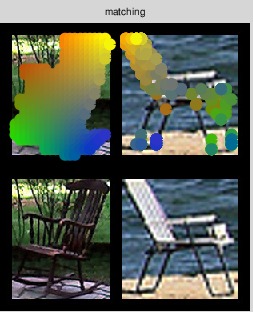} & 
\includegraphics[width=0.30\linewidth]{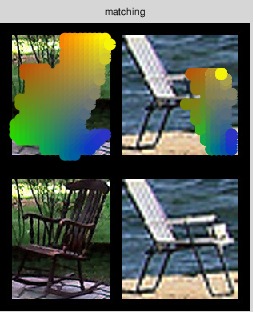} \\
\end{tabular}
\caption{\label{fig:alignments} Example chair alignments. On the left the proposed method VVN, on the right SIFTflow.}
\end{figure*}


\begin{figure*}
\centering
\renewcommand{\arraystretch}{1}
\begin{tabular}{@{}c@{} c@{}}
\includegraphics[width=0.49\linewidth]{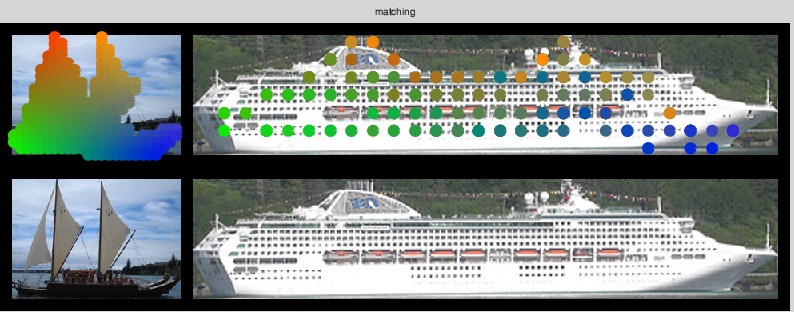} &
\includegraphics[width=0.49\linewidth]{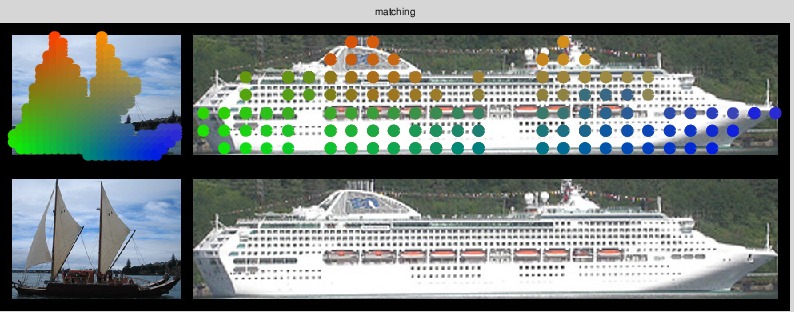} \\
\includegraphics[width=0.49\linewidth]{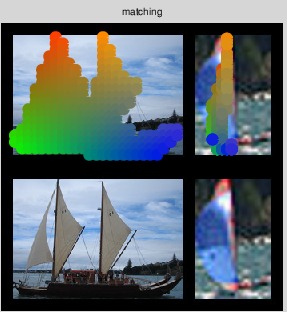} &
\includegraphics[width=0.49\linewidth]{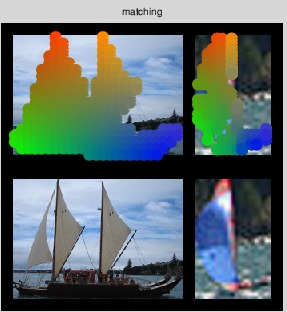} \\ 
\includegraphics[width=0.49\linewidth]{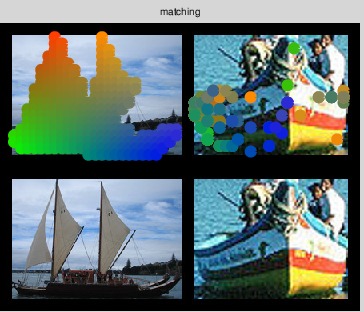} & 
\includegraphics[width=0.49\linewidth]{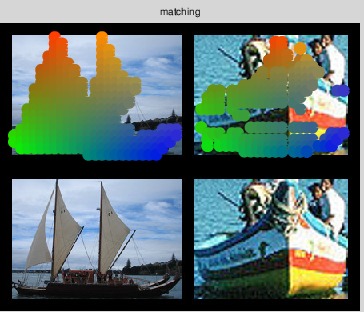} \\
\end{tabular}
\caption{\label{fig:alignments} Example boat alignments. On the left the proposed method VVN, on the right SIFTflow.}
\end{figure*}


\begin{figure*}
\centering
\renewcommand{\arraystretch}{1}
\begin{tabular}{@{}c@{} c@{}}
\includegraphics[width=0.35\linewidth]{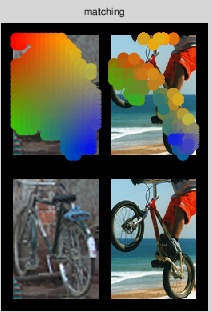} &
\includegraphics[width=0.35\linewidth]{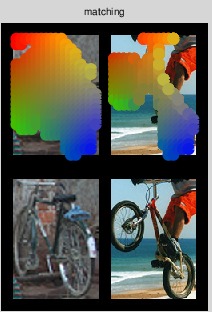} \\
\includegraphics[width=0.35\linewidth]{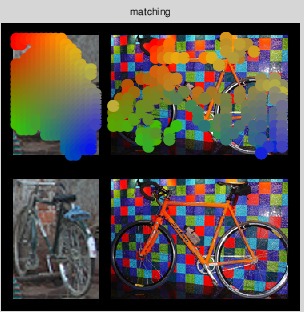} &
\includegraphics[width=0.35\linewidth]{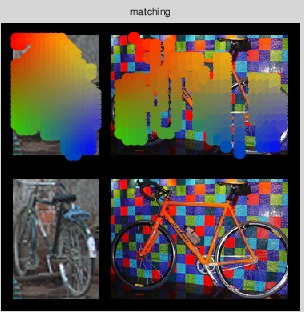} \\ 
\includegraphics[width=0.35\linewidth]{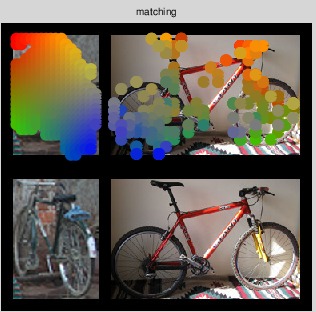} & 
\includegraphics[width=0.35\linewidth]{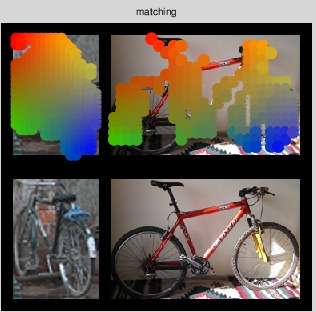} \\
\end{tabular}
\caption{\label{fig:alignments} Example bicycle alignments. On the left the proposed method VVN, on the right SIFTflow.}
\end{figure*}


\begin{figure*}
\centering
\renewcommand{\arraystretch}{1}
\begin{tabular}{@{}c@{} c@{}}
\includegraphics[width=0.49\linewidth]{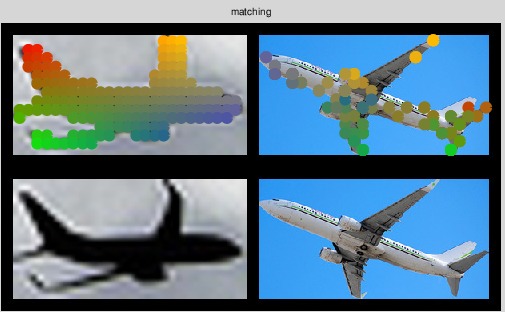} &
\includegraphics[width=0.49\linewidth]{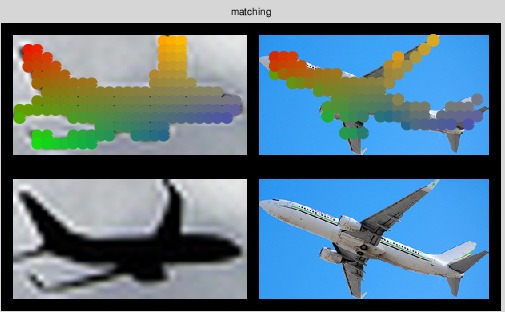} \\
\includegraphics[width=0.49\linewidth]{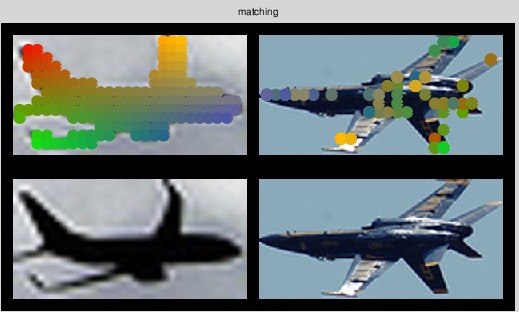} &
\includegraphics[width=0.49\linewidth]{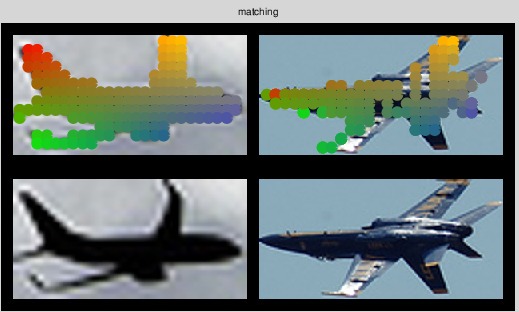} \\ 
\includegraphics[width=0.49\linewidth]{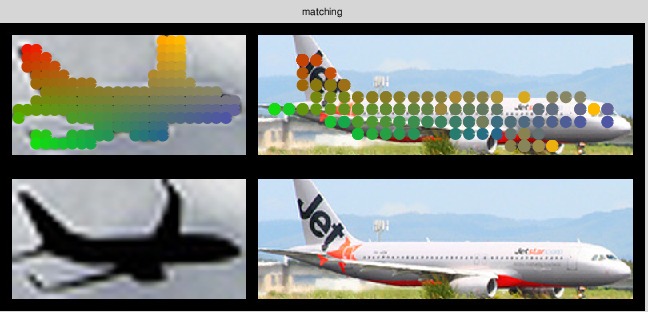} & 
\includegraphics[width=0.49\linewidth]{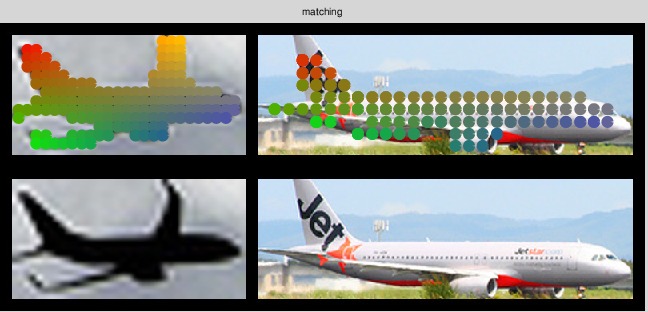} \\
\end{tabular}
\caption{\label{fig:alignments} Example aeroplane alignments. On the left the proposed method VVN, on the right SIFTflow.}
\end{figure*}


\end{document}